%% file: main.tex
\documentclass[lettersize,journal]{IEEEtran}
\usepackage{amsmath,amsfonts}
\usepackage{algorithm}
\usepackage{algpseudocode} 
\usepackage{array}
\usepackage[caption=false,font=normalsize,labelfont=sf,textfont=sf]{subfig}
\usepackage{textcomp}
\usepackage{stfloats}
\usepackage{url}
\usepackage{verbatim}
\usepackage{graphicx}
\usepackage{cite}
\usepackage{adjustbox}
\usepackage{enumitem}
\usepackage{booktabs}
\usepackage{multirow}
\usepackage{xcolor}
\usepackage{siunitx}
\usepackage{caption} 
\usepackage{colortbl}
\begin{document}

\title{FedSKD: Aggregation-free Model-heterogeneous Federated Learning via Multi-dimensional Similarity Knowledge Distillation for Medical Image Classification}



\author{Ziqiao~Weng,
        Weidong~Cai,
        and~Bo~Zhou%
\thanks{Ziqiao Weng and Bo Zhou are with the Department of Radiology,
Northwestern University, Chicago, IL, USA. 
Bo Zhou is the corresponding author (e-mail: bo.zhou@northwestern.edu).}

\thanks{Ziqiao Weng and Weidong Cai are with the School of Computer Science,
The University of Sydney, Sydney, NSW, Australia.}
}

\markboth{Journal of \LaTeX\ Class Files,~Vol.~14, No.~8, August~2021}%
{Shell \MakeLowercase{\textit{et al.}}: A Sample Article Using IEEEtran.cls for IEEE Journals}


\maketitle

\begin{abstract}

Federated learning (FL) enables privacy-preserving collaborative model training without direct data sharing. Model-heterogeneous FL (MHFL) enables clients to train personalized models with heterogeneous architectures, but existing methods mainly rely on centralized aggregation or require partially identical architectures, limiting scalability and efficiency. Current peer-to-peer (P2P) FL frameworks, though removing server dependence, have not been adapted to heterogeneous models and suffer from model drift and knowledge dilution. To address these challenges, we propose FedSKD, a novel P2P MHFL framework for medical image classification that facilitates direct knowledge exchange through round-robin model circulation, eliminating the need for centralized aggregation while allowing fully heterogeneous model architectures across clients. FedSKD’s key innovation lies in multi-dimensional similarity knowledge distillation, which enables bidirectional cross-client knowledge transfer at batch, pixel/voxel, and region levels for heterogeneous models in FL. This approach mitigates catastrophic forgetting and model drift through progressive reinforcement and distribution alignment while preserving model heterogeneity. Extensive evaluations on fMRI-based autism spectrum disorder diagnosis and skin lesion classification demonstrate that FedSKD outperforms state-of-the-art heterogeneous and homogeneous FL baselines, achieving superior personalization and cross-institutional generalization. These findings underscore FedSKD’s potential as a scalable and robust solution for real-world medical federated learning.

\end{abstract}


\input{content/introduction}

\input{content/related_work}

\input{content/method}

\input{content/experiment}

\input{content/discussion}

\input{content/conclusion}

\section*{Acknowledgments}

This research was supported by Australian Government Research Training Program (RTP) scholarship.

\bibliographystyle{IEEEtran}
\bibliography{main}

\clearpage
\appendix
\section*{Supplementary Material}
\input{supplementary}

\vfill

\end{document}

%% file: content/introduction.tex
\section{Introduction}\label{sec:introduction}

\IEEEPARstart{A}{rtificial} intelligence (AI)-based medical image analysis is fundamentally constrained by limited annotated data and strict privacy regulations. Medical imaging data are costly to acquire and annotate, while regulations such as HIPAA \cite{kels2020hipaa}, GDPR \cite{voigt2017eu}, and the Cyber Security Law \cite{parasol2018impact} severely restrict centralized data sharing across institutions. These challenges motivate learning paradigms that can leverage multi-institutional data without direct data aggregation.

\begin{figure}[t]
\begin{minipage}[b]{1.0\linewidth}
    \centering
    \centerline{\includegraphics[width=\columnwidth]{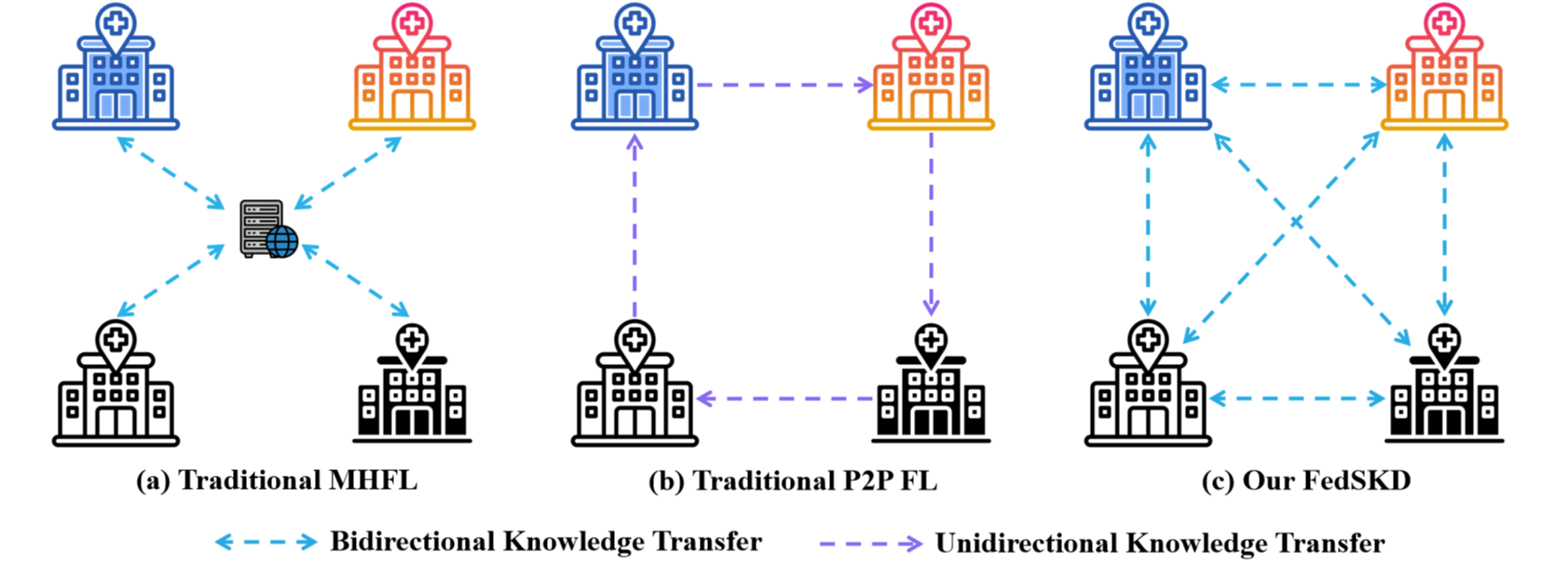}}
    
 \caption{\small{Comparison of traditional MHFL schemes (a), traditional sequential P2P FL schemes (b), and our proposed MHFL framework (c). Traditional MHFL methods rely on a central server for knowledge fusion and require partially identical model structures across clients. Traditional P2P FL trains a homogeneous model sequentially across clients. In contrast, our FedSKD framework operates without a central server and allows fully heterogeneous model architectures while enabling direct knowledge transfer among clients.}}
	\label{fig:fig1}
	\vspace{-15px}
 \end{minipage}
 \vspace{-10px}
\end{figure}

Federated learning (FL) enables collaborative model training across multiple institutions without direct data sharing \cite{li2020multi,peng2022fedni,fan2021federated}. Existing FL paradigms can be broadly categorized into central server-based and peer-to-peer approaches. In server-based FL, data remain decentralized at local institutions, while a centralized server coordinates the iterative optimization of a shared global model. The widely adopted FedAvg algorithm \cite{mcmahan2017communication} updates the global model by averaging client model weights; however, its naive aggregation strategy often performs poorly under non-IID data distributions, which are prevalent in medical imaging due to heterogeneous acquisition protocols, scanner types, and patient demographics \cite{yan2023label,wicaksana2022customized}. Such distribution mismatches can impair convergence and generalization \cite{guan2024federated,yashwanth2023adaptive}. To mitigate these issues, several extensions have been proposed, including FedProx \cite{li2020federated}, FedBN \cite{li2021fedbn}, as well as approaches based on domain adaptation \cite{li2020multi,dinsdale2022fedharmony,guo2021multi} and knowledge distillation \cite{he2023dealing,yashwanth2023adaptive,zhu2021data,gong2021ensemble}.

Peer-to-peer federated learning (P2P FL) \cite{tom2017designing,rieke2020future,10715722} removes the centralized server and enables direct model exchange among clients. As in server-based FL, non-IID data remains a fundamental challenge. However, classic P2P FL typically adopts a sequential learning protocol, where a single model is iteratively fine-tuned across clients rather than synchronously aggregated. This sequential exposure to heterogeneous data distributions causes the model to repeatedly adapt to shifting optimization objectives, leading to pronounced parameter oscillations and unstable convergence \cite{criado2022non}, a phenomenon commonly referred to as \emph{model drift}. Moreover, as the model progressively adapts to each new client, knowledge acquired from earlier clients is susceptible to interference or overwriting, resulting in a gradual erosion of previously learned representations, which we denote as \emph{knowledge dilution}. While other challenges in decentralized FL, such as communication overhead, computational redundancy, and privacy risks, have been extensively studied and partially mitigated \cite{kairouz2021advances,shi2021over,chen2022decentralized,roy2019braintorrent,yuan2023peer,xu2024federated}, the absence of effective mechanisms to address sequential model drift and knowledge dilution remains a key obstacle to robust convergence and sustained performance in P2P FL.

While data heterogeneity has been widely studied in FL, model heterogeneity remains a critical yet underexplored challenge. Variations in clients' computational resources and application-specific requirements often necessitate the use of heterogeneous model architectures \cite{li2019fedmd, afonin2021towards, huang2024overcoming}. Existing model-heterogeneous FL (MHFL) methods rely heavily on centralized aggregation (Figure \ref{fig:fig1} (a)) and generally follow two strategies, representation-level knowledge transfer \cite{li2019fedmd,wang2024towards,10373104} or partial weight aggregation \cite{wu2022communication,yi2024fedmoe,yi2024pfedafm}, but both face limitations such as reliance on public data, restricted architectural flexibility, vulnerability to gradient conflicts under non-IID settings, and scalability constraints imposed by the central server. In contrast, P2P FL enables collaboration among fully heterogeneous client models without server-side aggregation, offering a more flexible and scalable alternative.

However, existing P2P FL frameworks still suffer from model drift and knowledge dilution. Furthermore, whether based on parameter aggregation or alternative synchronization strategies, they generally assume a single homogeneous model that is sequentially trained across clients, as illustrated in Figure \ref{fig:fig1} (b). This assumption limits the flexibility of P2P frameworks in model-heterogeneous scenarios, where clients often use customized architectures tailored to their specific needs.

To address these issues, we propose FedSKD, a novel model-heterogeneous P2P FL framework based on multi-dimensional similarity knowledge distillation (SKD). FedSKD introduces two key innovations. First, it enables decentralized knowledge transfer by circulating locally trained heterogeneous models among clients in a round-robin manner, eliminating the gradient conflicts and server overhead inherent in traditional MHFL methods, as shown in Figure \ref{fig:fig1} (c). In each training round, each client receives a model from the previous client, performs bidirectional knowledge distillation, and passes its enhanced model to the next client. This closed-loop knowledge reinforcement chain ensures that critical features are progressively amplified, rather than diluted, counteracting the problem of knowledge dilution. Second, FedSKD incorporates multi-dimensional similarity knowledge distillation to facilitate efficient cross-client knowledge reinforcement among heterogeneous models. Specifically, we introduce three novel SKD mechanisms that operate purely at the feature representation level: Batch-wise SKD (B-SKD) to align activation patterns across batches, Pixel/Voxel-wise SKD (P-SKD) to align spatial feature distributions, and Region-wise SKD (R-SKD) to align representations within anatomical regions. During training, clients simultaneously distill domain-specific knowledge into received models while absorbing cross-institutional insights from these models, creating a bidirectional knowledge exchange that facilitates knowledge reinforcement and prevents model drift through continuous distribution alignment. 
We evaluate FedSKD on two clinically significant medical imaging tasks: fMRI-based autism spectrum disorder (ASD) diagnosis and skin lesion classification. Experimental results demonstrate that FedSKD outperforms existing methods in both personalization (client-specific performance) and generalization (cross-institutional adaptability), while addressing the challenges outlined above. These findings highlight FedSKD's potential to significantly advance federated learning in resource-constrained, need-variant, privacy-sensitive medical applications. Our key contributions can be summarized as follows:
\begin{itemize}
    \item We propose the first peer-to-peer model-heterogeneous federated learning (P2P-MHFL) framework tailored for medical image classification, enabling clients with diverse model architectures to collaboratively learn without relying on a central server. 
    \item Beyond extending traditional sequential P2P FL into the model-heterogeneous setting, our method introduces a novel multi-dimensional similarity knowledge distillation mechanism, which effectively mitigates two fundamental challenges unique to P2P FL: \textit{model drift} and \textit{knowledge dilution}.
    \item We construct a geographically partitioned federated dataset based on ABIDE, representing the first such dataset in the literature, and demonstrate that our method achieves state-of-the-art performance on two carefully curated medical benchmarks, FedASD and FedSkin. This dataset can also serve as a valuable resource for future research. 
\end{itemize}

%% file: content/related_work.tex
\section{Related Work}

\subsection{Model-Heterogeneous Federated Learning}

Beyond data heterogeneity, model heterogeneity has emerged as a critical challenge in federated learning, driven by differences in clients’ computational capacities and task-specific architectural requirements. Existing MHFL approaches predominantly rely on centralized aggregation and can be broadly categorized into two groups.

Representation aggregation–based MHFL:
These methods exchange knowledge through shared representations, often derived from public datasets \cite{li2019fedmd,chang2019cronus,lin2020ensemble,cheng2021fedgems,itahara2021distillation,cho2022heterogeneous,wang2024towards}. For instance, FedMD \cite{li2019fedmd} constructs class-wise representations from public data and distills them to heterogeneous client models. However, acquiring suitable public datasets is often infeasible in privacy-sensitive medical domains. To address this limitation, AlFeCo \cite{10373104} introduces an allosteric feature generator that enables task-relevant representation exchange across heterogeneous models without relying on public data.

Weight aggregation–based MHFL:
These approaches partially align heterogeneous architectures by co-training client-specific models with auxiliary homogeneous subnetworks or experts \cite{wu2022communication,yi2023fedssa,qin2023fedapen,shen2023federated,yi2024federated,yi2024fedmoe,yi2024pfedafm}. FedMoE \cite{yi2024fedmoe} assigns each client a local heterogeneous expert while sharing a global expert, combining their outputs through a gating mechanism. pFedAFM \cite{yi2024pfedafm} further balances features using a learnable adaptive vector. Despite their effectiveness, these methods remain susceptible to gradient conflicts under non-IID data \cite{xu2024federated}, and full architectural heterogeneity is constrained since portions of the model must remain shared. Moreover, reliance on centralized aggregation introduces scalability limitations as well as communication and storage overhead.

\subsection{Peer-to-Peer Federated Learning}

Peer-to-Peer Federated Learning (P2P FL), also known as Decentralized Federated Learning (DFL), enables clients to directly exchange knowledge without a centralized server for aggregation or coordination \cite{lalitha2019peer,yuan2023peer}. In this paradigm, the communication topology plays a central role by governing information exchange and fusion across clients. Classical topologies include line, ring, mesh, star, and tree, each offering distinct trade-offs in communication cost, computational load, convergence behavior, and robustness. More recently, hybrid topologies have attracted increasing attention by combining the strengths of multiple structures to better accommodate diverse system constraints \cite{yuan2024decentralized}.

Representative work includes the hybrid P2P FL network proposed by Xing et al. \cite{xing2020decentralized}, which employs neighbor-restricted broadcast gossip to improve resilience in unreliable wireless networks. Building on this design, Shi et al. \cite{shi2021over} incorporated coding schemes, gradient tracking, and variance reduction to further enhance convergence. Wang et al. \cite{wang2022matcha} proposed Matcha, a dynamic DFL framework that adaptively modifies the topology during training. By prioritizing critical communication links, Matcha accelerates convergence while reducing latency and communication overhead.

From the perspective of knowledge integration, P2P FL can be divided into two paradigms: Aggregate and Continual. In the Aggregate paradigm \cite{roy2019braintorrent,pappas2021ipls,chen2022decentralized}, each client aggregates models received from multiple peers before local training, enabling richer knowledge fusion at the cost of increased communication, computation, and storage. In contrast, the Continual paradigm \cite{sheller2020federated,yoon2021federated,huang2022continual} restricts each client to receiving a single peer model at a time and directly updating it via local learning, significantly simplifying system overhead. However, convergence is highly sensitive to the propagation order and often requires more communication rounds to reach consensus \cite{yuan2024decentralized}. As the model sequentially traverses clients with heterogeneous data, catastrophic forgetting and concept drift become prominent challenges that undermine training stability \cite{criado2022non}. In contrast, our proposed FedSKD extends P2P FL to model-heterogeneous settings and mitigates these issues through a similarity-based knowledge distillation mechanism with round-robin circulation.

%% file: content/method.tex
\section{Method}

\subsection{Problem Definition}
In this study, we investigate MHFL for supervised classification tasks in medical imaging. MHFL involves $N$ clients where each FL client $i$ possesses a local labeled dataset $D_i$ and a unique local model $M_i$, defined by a specific architecture $M_i$ and parameters $\theta_i$. Unlike traditional approaches, the local models can be heterogeneous across clients, such that $M_i \neq M_j, i,j\in\{1,...,N\}$ for different clients $i$ and $j$. All clients are assumed to perform the same classification task using data of the same modality. However, in real-world medical scenarios, the data distribution $P_i$ of each client’s dataset $D_i$ is typically non-IID. This creates dual challenges: managing both model heterogeneity and data heterogeneity, which makes MHFL significantly more complex than conventional model-homogeneous federated learning. The goal of MHFL is to minimize the aggregated loss across all clients’ heterogeneous local models with:
\begin{equation}
\scalebox{0.9}{$\displaystyle
\min\limits_{\theta_1,\ldots,\theta_{N}}\sum_{i=1}^{N}\mathcal{L}(M_i(D_{i})).
$}
\end{equation}
Our objective in this work is to train robust heterogeneous local models for each client through collaborative learning among decentralized clients without data sharing.

\subsection{Overall Framework of FedSKD}

Figure \ref{fig:fig2} (b) provides an overview of the training process for our FedSKD framework, illustrating the steps followed during each communication round of the training:

\begin{figure*}[t]
\begin{minipage}[b]{1.0\linewidth}
    \centering
    \centerline{\includegraphics[width=0.90\textwidth]{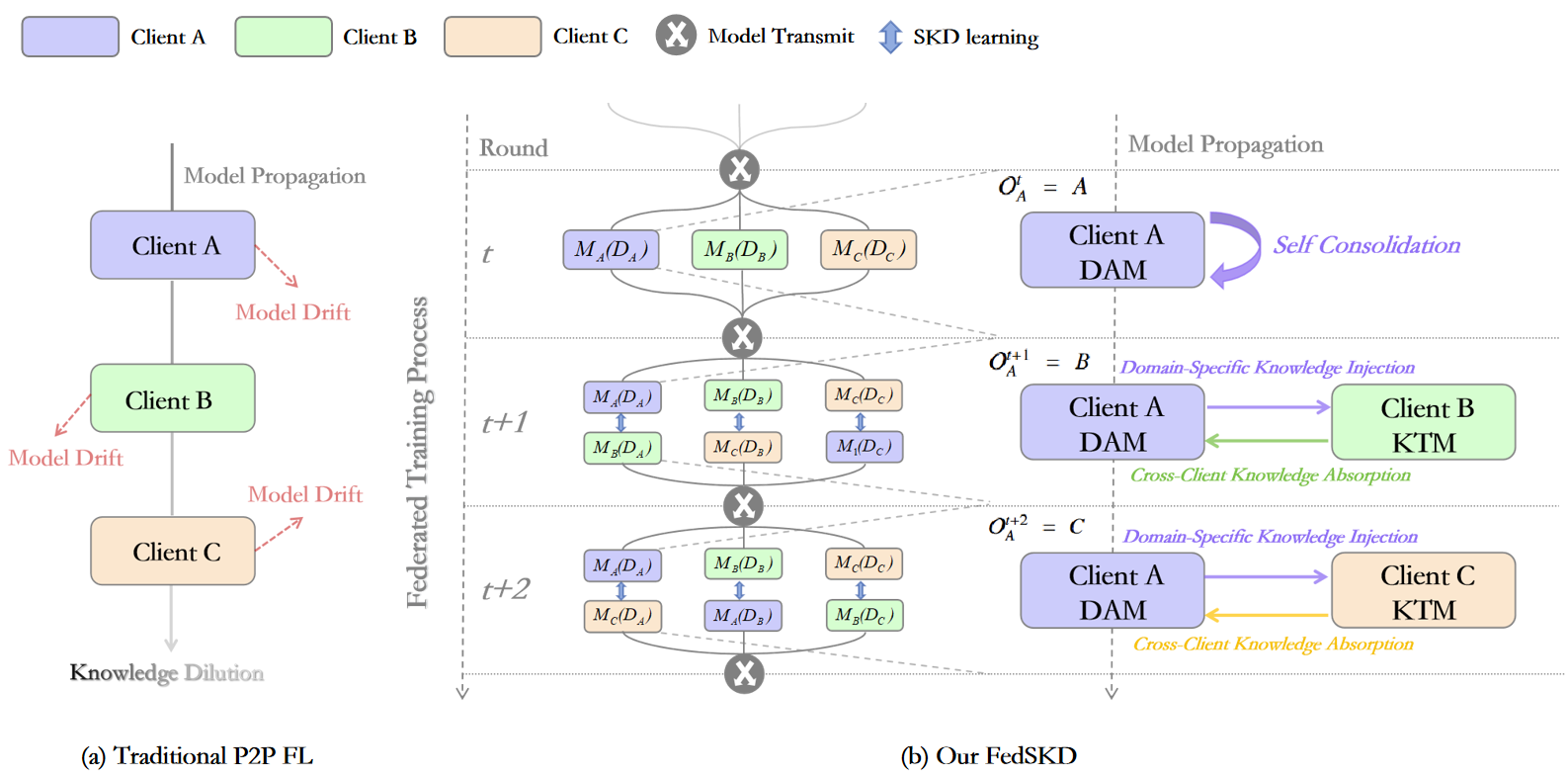}}

 \caption{\small{(a) Traditional P2P FL methods are prone to model drift and knowledge dilution, limiting their effectiveness in MHFL. (b) Our FedSKD framework enhances knowledge transfer by circulating heterogeneous models among clients in a round-robin manner guided by a predefined order $\mathcal{O}$. During non-transfer phases, clients refine their personalized domain-adaptive models (DAM) locally. In transfer phases, bidirectional knowledge transfer occurs between the local DAM and the received knowledge-transit model (KTM) through multi-dimensional similarity knowledge distillation (SKD). This process allows the DAM to share domain-specific knowledge with the KTM, while the KTM provides cross-client insights to the DAM, ensuring continuous knowledge reinforcement and mitigating model drift through consistent distribution alignment.}}
	\label{fig:fig2}
	\vspace{-15px}
 \end{minipage}
 \vspace{-10px}
\end{figure*}

\textcircled{1} Cross-Client Model Transfer Path Generation: First, we determine a random route for transferring each client's local model across clients. In each training round, every client receives a model transferred from a randomly assigned, non-repeating client, ensuring a one-to-one mapping between models and clients. We define the order of the \( t\)th round as \(  \mathcal{O}^{t}= \left \{  \mathcal{O}_{1}^{t}, ...,\mathcal{O}_{N}^{t} \right \} = \pi([1,2,\dots,N]) \), where \( \pi(\cdot)\) represents the random permutation operation and \( \mathcal{O}_{i}^{t}\) represents the client index from which client \( i \) receives the cross-client model. 

In this framework, we refer to the client's local model as the \textbf{Domain-Adaptive Model (DAM)}, which specializes in adapting to the local data distribution, and the received model as the \textbf{Knowledge-Transit Model (KTM)}, which serves as a carrier of cross-client knowledge. This structured method ensures efficient and unbiased knowledge propagation while maintaining the individuality of each client's model.

\textcircled{2} Bidirectional Knowledge Distillation: At the beginning of each round, each client receives a copy of a latest model transmitted from its assigned remote client according to \( \mathcal{O}^{t} \). Taking client \( i \) as an example, if \( \mathcal{O}_{i}^{t} = i \), the client \( i \) performs self consolidation, locally refining its DAM solely on \( D_i \) through standard supervised learning. Conversely, if \( \mathcal{O}_{i}^{t} = j \), \( (i \ne j) \), client \( i \) performs bidirectional knowledge distillation between its local DAM $M_i$ and the received KTM \(\tilde{M_j}\).

The distillation progresses through two concurrent knowledge flows: \textbf{Domain-Specific Knowledge Injection (DAM $\rightarrow$ KTM)}: client \( i \)'s DAM \(M_i\) acts as a local expert, distilling its domain-specific expertise from $D_i$ into the received KTM \(\tilde{M_j}\) through multi-dimensional SKD. \textbf{Cross-Client Knowledge Absorption (KTM $\rightarrow$ DAM)}: Simultaneously, The client \( j \)'s KTM \(\tilde{M_j}\) serves as a carrier of cross-client knowledge, providing complementary insights to enhance the generalization capability of the client \( i \)'s DAM \(M_i\). In addition to mutual distillation, client \( i \)'s DAM \(M_i\) receives direct supervision from \( D_j \) to facilitate local adaptation and client \( j \)'s KTM \(\tilde{M_j}\) continues local training on \( D_j \) to maintain base performance and avoid severe model drift. This dual-phase process guarantees progressive knowledge refinement as received KTMs adapt to local distributions while retaining cross-client knowledge, and local DAMs assimilate refined knowledge through reciprocal distillation. Catastrophic forgetting (i.e. knowledge dilution) is mitigated via multi-dimensional SKD's feature-space regularization (detailed in Section \ref{sec:skd}).

\textcircled{3} Model Decomposition and Knowledge Preservation: We decompose the model into two functionally distinct components: a feature extractor, which transforms raw input data into high-level feature representations, and a prediction header, which serves as a classifier to generate final predictions based on these features. Our primary objective is to ensure that, during adaptation to the target data distribution, the received model retains its previously acquired cross-client knowledge, rather than allowing the target domain's knowledge to progressively overwrite it. To achieve this, we freeze the parameters of the received model's prediction header and only finetuning the feature extractor during mutual learning. This constraint serves two critical purposes: 1) it prevents the feature extractor from over-adapting to the target domain, thereby preserving the cross-client knowledge embedded in the model, and 2) it regularizes the shift in feature representations from the received model's domain to the target domain, ensuring stable and generalizable feature alignment. Moreover, while the local feature extractor in DAM assimilates cross-client knowledge from the transferred KTM, the local prediction head is exclusively optimized on client-specific data. This design ensures that each client maintains its unique specialization while mitigating the risk of the feature extractor excessively adapting to external domains. Taken together, these mechanisms facilitate the emergence of a consensus feature representation across clients, while simultaneously preserving local expertise.

\textcircled{4} Model Update and Propagation: Upon completing the training in the \( t\)-th round, each client discards the received model (KTM) and retains only its updated and enhanced local model (DAM). The updated models are then propagated to new clients in the subsequent round according to the permutation order \( \mathcal{O}^{t+1} \), ensuring continuous and structured knowledge exchange across the network. This iterative process enables progressive model refinement while maintaining decentralized coordination.

\textcircled{5} Joint Optimization Objective: The total loss of SKD training at the \( t\)-th stage on a client can be defined as follows:

\begin{equation} 
\scalebox{0.9}{$\displaystyle
\begin{split}
\mathcal{L}(M_i^t,\tilde{M_j^t})&=\mathcal{L}_{\text{CE}}(M_i^t(x_i), y_i) \\ 
&\quad + \gamma \cdot \mathcal{L}_{\text{SKD}}(M_i^t(x_i), \tilde{M_j^t}(x_i)) \\
&\quad + \mathcal{L}_{\text{CE}}(\tilde{M_j^t}(x_i), y_i)
\end{split}
\label{eq:eq2}
$}
\end{equation}
where \(\mathcal{L}_{\text{CE}}\) and \( \mathcal{L}_{\text{SKD}}\) denotes the cross-entropy loss and our multi-dimensional similarity knowledge distillation loss (to be elaborated in Section \ref{sec:skd}). Here, \(\gamma\) is a balancing hyperparameter, and \( (x_i,y_i) \sim D_i\).

The global optimization objective during this stage thus can be formulated as:

\begin{equation} 
\scalebox{0.9}{$\displaystyle\min\limits_{\theta_1,\ldots,\theta_{N}}\sum_{i=1}^{N}\mathcal{L}(M_i^{t},\tilde{M_{\mathcal{O}_{i}^{t}}}). 
$}
\end{equation}

The above steps are iteratively performed until the local models of all clients have converged across the entire set of client datasets. Subsequently, each client retains its respective model for the purposes of federated learning deployment and inference. A simplified version of the FedSKD workflow is provided in Algorithm 1 in the Supplementary Material.

\subsection{Multi-dimensional Similarity Knowledge Distillation} \label{sec:skd}

\begin{figure*}[t]
\begin{minipage}[b]{1.0\linewidth}
    \centering
    \centerline{\includegraphics[width=0.9\textwidth]{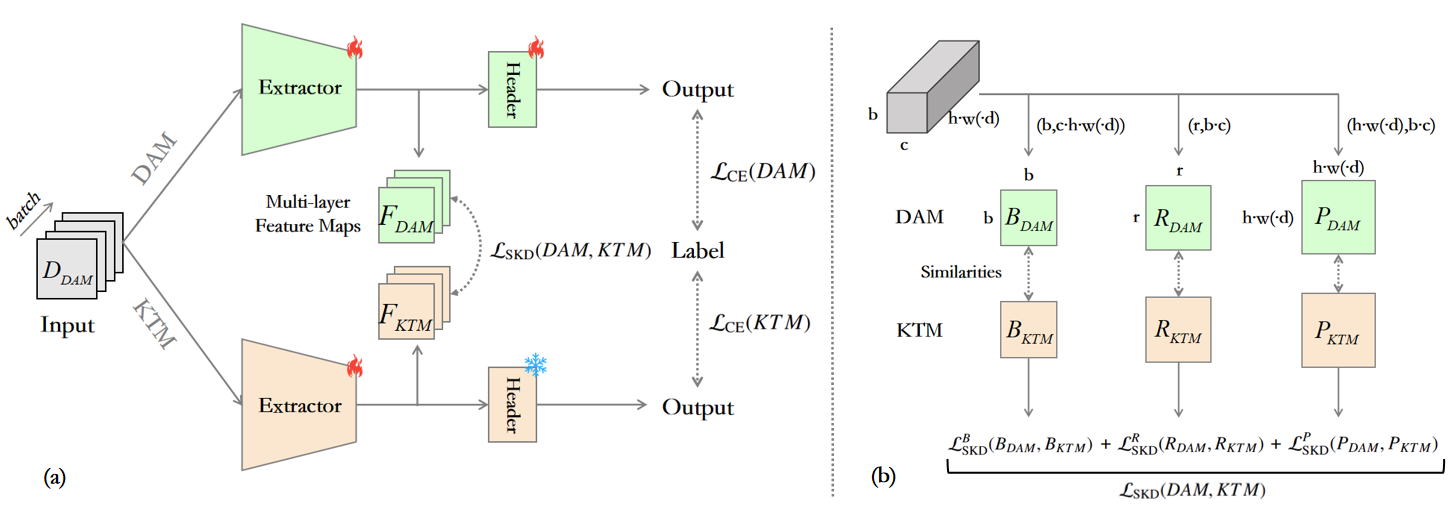}}
    
 \caption{\small{(a) Overview of the FedSKD training framework: Multi-dimensional similarity knowledge distillation facilitates the mutual distillation of semantically meaningful knowledge between the local DAM and the received KTM. Both models are jointly optimized on the DAM's dataset, leveraging supervised learning with corresponding dataset labels. The snowflake icon signifies frozen parameters, whereas the flame icon represents active parameter updates during training. (b) Computation of multi-dimensional similarity knowledge distillation loss: The loss is calculated by measuring divergence between the learned similarity patterns of the DAM and KTM across multiple granularities (Batch-wise, Pixel/Voxel-wise, Region-wise), ensuring effective knowledge transfer and alignment.}}
	\label{fig:fig3}
	\vspace{-15px}
 \end{minipage}
 \vspace{-10px}
\end{figure*}

We propose a multi-dimensional similarity knowledge distillation approach within the context of our model-heterogeneous federated learning (FL) framework (Figure \ref{fig:fig3}). The specifics of this approach are detailed below:

\textbf{Notation}: Without loss of generality, we consider a 2D scenario here. Given a mini-batch input, denote the feature map produced by the local model $DAM$ at a particular layer $\ell$ by $F_{DAM}^{\ell}\in\mathbb{R}^{ b\times c\times h\times w}$, and the feature map produced by the received network $KTM$ at a particular layer $\ell^{\prime}$ by $F_{KTM}^{\ell^{\prime}}\in\mathbb{R}^{ b\times c^{\prime} \times h^{\prime} \times w^{\prime}}$, where \(b\) is the batch size, \(c\) is the channel number, and \(h\) and \(w\) are spatial dimensions. Note that \(c', h', w'\) do not necessarily have to be identical to \(c, h, w\), as the models are heterogeneous. In our setting, since the local network and the received network have the same depth (i.e., the number of blocks), we let \(\ell = \ell^{\prime}\), meaning that we perform knowledge distillation at the same layer for both the local model $DAM$ and the received model $KTM$. We denote \(\mathbb{L}\) as the set of all \( \ell \) in knowledge distillation process.

\textbf{Batch-wise Similarity} quantifies inter-sample semantic relationships across categories within a mini-batch. The batch-wise similarity loss guides the local model $DAM$ and the received model $KTM$ towards capturing  identical dataset-specific semantic patterns. We first reshape the feature map \(F_{DAM}^{\ell} \) of the DAM to \( H_{DAM}^{\ell}\in\mathbb{R}^{ b\times chw} \). We then compute the batch-wise similarity induced in the DAM's feature map \(F_{DAM}^{\ell} \) as follows: 
\begin{equation} 
\scalebox{0.9}{$\displaystyle
B_{DAM}^{\ell}=\frac{H_{DAM}^{\ell} \cdot (H_{DAM}^{\ell})^{T}}{\left \|  H_{DAM}^{\ell} \cdot (H_{DAM}^{\ell})^{T} \right \|_{2}^{(1)} } \cdot \sqrt{b} 
$}
\end{equation}
where \( B_{DAM}^{\ell}\) is a \( b \times b\) matrix, and \( \left \| \cdot  \right \|_{2}^{(1)} \) is the row-wise L2 normalization (i.e. in dim=1 of the matrix). Each item in the matrix is indirectly divided by \( \sqrt{b}\) during the computation, so we multiply by  \( \sqrt{b}\) at the end to scale the values back, bringing them to the same magnitude as other forms of similarity. Intuitively, the entry \( (i, j) \) in \( B_{T}^{\ell} \) represents the scaled similarity activated at layer \( l\) of the DAM between the $i$-th and $j$-th images in the mini-batch. Analogously, the batch-wise similarity in the KTM can be calculated by:

\begin{equation} 
\scalebox{0.9}{$\displaystyle
B_{KTM}^{\ell}=\frac{H_{KTM}^{\ell} \cdot (H_{KTM}^{\ell})^{T}}{\left \|  H_{KTM}^{\ell} \cdot (H_{KTM}^{\ell})^{T} \right \|_{2}^{(1)} } \cdot \sqrt{b} 
$}
\end{equation}

Then, we define the batch-wise similarity knowledge distillation loss as:
\begin{equation} 
\scalebox{0.9}{$\displaystyle
\mathcal{L}_{\text{SKD}}^{B}(B_{DAM},B_{KTM})= \frac{1}{\left | \mathbb{L} \right | } \sum_{\ell \in \mathbb{L} }^{}\frac{1}{b^{2}}\left \| B_{DAM}^{\ell} - B_{KTM}^{\ell} \right \|_{F}^{2}   
$}
\end{equation}

\textbf{Pixel/Voxel-wise Similarity} measures fine-grained semantic correspondences at individual spatial units (pixels/voxels). The pixel/voxel-wise similarity loss enforces spatial consistency between the feature representations of the $DAM$ and the $KTM$, ensuring that both models align in their understanding of local structural patterns and spatial relationships within a mini-batch. We first reshape the feature map \(F_{DAM}^{\ell} \) of the DAM to \( H_{DAM}^{\ell}\in\mathbb{R}^{ hw\times bc} \). We then compute the pixel-wise similarity induced in the DAM's feature map \(F_{DAM}^{\ell} \) as follows: 
\begin{equation} 
\scalebox{0.9}{$\displaystyle
P_{DAM}^{\ell}=\frac{H_{DAM}^{\ell} \cdot (H_{DAM}^{\ell})^{T}}{\left \|  H_{DAM}^{\ell} \cdot (H_{DAM}^{\ell})^{T} \right \|_{2}^{(1)} } \cdot \sqrt{hw} 
$}
\end{equation}

Analogously, the pixel-wise similarity in the KTM can be calculated by:

\begin{equation} 
\scalebox{0.9}{$\displaystyle
P_{KTM}^{\ell}=\frac{H_{KTM}^{\ell} \cdot (H_{KTM}^{\ell})^{T}}{\left \|  H_{KTM}^{\ell} \cdot (H_{KTM}^{\ell})^{T} \right \|_{2}^{(1)} } \cdot \sqrt{hw} 
$}
\end{equation}

Then, we can define the pixel-wise similarity knowledge distillation loss as:
\begin{equation} 
\scalebox{0.9}{$\displaystyle
\mathcal{L}_{\text{SKD}}^{P}(P_{DAM},P_{KTM})= \frac{1}{\left | \mathbb{L} \right | } \sum_{\ell \in \mathbb{L} }^{}\frac{1}{hw^{2}}\left \| P_{DAM}^{\ell} - P_{KTM}^{\ell} \right \|_{F}^{2} 
$}
\end{equation}

Given the potential structural heterogeneity of models, it is necessary to align $P_{DAM}$ and $P_{KTM}$ to the same spatial dimension before computing the loss, especially when their spatial dimensions are mismatched. This dimension-matching step can be straightforwardly achieved using various spatial interpolation techniques, such as bilinear interpolation.

\textbf{Region-wise Similarity} captures semantic correlations between meaningful, interrelated regions within a given modality. The region-wise similarity loss enforces consistency between the DAM and the KTM, guiding both models to learn similar inter-region correlations within a mini-batch. For instance, in neuroimaging, the brain is partitioned into functional regions, which exhibit strong interdependencies despite anatomical separation \cite{friston2011functional,bullmore2009complex}. Such region-based correlations are crucial for understanding complex systems, making region-wise similarity a powerful tool for capturing high-level relationships in diverse applications. Given an image with \( r \) predefined regions, let \( M_{k} \) be the collection of all pixels \( (i,j) \) belonging to the $k$th region. We first compute the feature corresponding to the \( k \)th region by averaging the features within this region as follows:

\begin{equation} 
\scalebox{0.9}{$\displaystyle
H_k^{(b,c)} = \frac{1}{\left | M_{k} \right | } \sum_{(i,j) \in M_{k}} F^{(b,c,h,w)}(i,j)
$}
\end{equation}
Then, we can get the region-wise feature maps via:
\begin{equation} 
\scalebox{0.9}{$\displaystyle
H^{(b, r, c)} = \left[ H_1^{(b,c)}, H_2^{(b,c)}, \dots, H_r^{(b,c)} \right],
$}
\end{equation}
where the output can then be reshaped as \( H\in\mathbb{R}^{ r\times bc} \). Similarly, we compute the region-wise similarity induced in the DAM's feature map \(F_{DAM}^{\ell} \) and the KTM's feature map \(F_{KTM}^{\ell} \) as follows: 

\begin{equation} 
\scalebox{0.9}{$\displaystyle
R_{DAM}^{\ell}=\frac{H_{DAM}^{\ell} \cdot (H_{DAM}^{\ell})^{T}}{\left \|  H_{DAM}^{\ell} \cdot (H_{DAM}^{\ell})^{T} \right \|_{2}^{(1)} } \cdot \sqrt{r} ,
$}
\end{equation}

\begin{equation} 
\scalebox{0.9}{$\displaystyle
R_{KTM}^{\ell}=\frac{H_{KTM}^{\ell} \cdot (H_{KTM}^{\ell})^{T}}{\left \|  H_{KTM}^{\ell} \cdot (H_{KTM}^{\ell})^{T} \right \|_{2}^{(1)} } \cdot \sqrt{r} .
$}
\end{equation}

Given above, we define the region-wise similarity knowledge distillation loss as:
\begin{equation} 
\scalebox{0.9}{$\displaystyle
\mathcal{L}_{\text{SKD}}^{R}(R_{DAM},R_{KTM})= \frac{1}{\left | \mathbb{L} \right | } \sum_{\ell \in \mathbb{L} }^{}\frac{1}{r^{2}}\left \| R_{DAM}^{\ell} - R_{KTM}^{\ell} \right \|_{F}^{2} 
$}
\end{equation}

Combining all above, we then define the total loss of multi-dimensional similarity knowledge distillation as:

\begin{equation}
\scalebox{0.9}{$\displaystyle
\begin{split}
\mathcal{L}_{\text{SKD}}(DAM,KTM) &= \mathcal{L}_{\text{SKD}}^{B}(B_{DAM},B_{KTM}) \\
&\quad + \mathcal{L}_{\text{SKD}}^{P}(P_{DAM},P_{KTM}) \\
&\quad + \mathcal{L}_{\text{SKD}}^{R}(R_{DAM},R_{KTM})
\end{split}
$}
\end{equation}

Please note that if the dataset does not have predefined region masks, the region-wise similarity loss is optional.

%% file: content/experiment.tex
\section{Experiments and Results}
The experiments were conducted on two challenging classification tasks using real-world medical imaging datasets to assess the effectiveness of our methods. In the following sections, we detail our tasks and datasets (Section \ref{sec:task&data}), experimental setups (Section \ref{sec:setup}), and experimental results (Section \ref{sec:results}).

\subsection{Tasks and Datasets} \label{sec:task&data}
\subsubsection{fMRI-based Autism Spectrum Disorder Classification}
\textbf{Task Description}: Autism Spectrum Disorder (ASD) is a neurodevelopmental disorder characterized by impairments in social interaction, communication, learning, and behavior \cite{sharma2018autism}. Early diagnosis is crucial for enabling timely and effective interventions. Resting-state functional magnetic resonance imaging (rs-fMRI) has emerged as a widely used non-invasive tool for capturing brain connectivity patterns, making it a valuable resource for the early diagnosis of neurological disorders. In this study, we leverage rs-fMRI data to differentiate individuals with ASD from typical controls (TC), framing the problem as a binary classification task.

\textbf{Dataset Description}: We utilize the publicly available ABIDE-I dataset \cite{di2014autism}, which includes data from 1,112 subjects with 539 diagnosed with ASD and 573 healthy controls. The data is collected across 17 sites in North America and Europe, which encompasses T1-weighted structural brain images, fMRI scans, and detailed phenotypic information for each subject. Following rigorous quality control and the exclusion of incomplete or corrupted data, we retained a final dataset of 871 subjects (403 with ASD and 468 healthy controls). Preprocessing was performed using the Connectome Computation System (CCS) pipeline. Nuisance signal removal was then applied to mitigate confounding variations arising from head motion, physiological processes (e.g., heartbeat and respiration), and scanner drift. After nuisance regression, a bandpass filter (0.01–10 Hz) was applied, excluding global signal correction. Spatial normalization was subsequently conducted to align the brain images with the Montreal Neurological Institute (MNI) template, achieving a standardized resolution of \(3 \times 3 \times 3\) mm³. In this study, we adopt the methodology proposed in STO \cite{weng2025efficient}, utilizing 3D statistical derivatives that preserve the full spatial resolution \(61 \times 73 \times 61\) of the original 4D fMRI data as network input. These derivatives include Regional Homogeneity (ReHo), Degree Centrality (DC), Local Functional Connectivity Density (LFCD), and Voxel-Mirrored Homotopic Connectivity (VMHC). For further details on these derivatives, we refer readers to \cite{craddock2013towards}.

\textbf{Non-IID Dataset Construction}: The ABIDE dataset is aggregated from 17 distinct medical centers across geographically diverse regions, with each institution employing distinct assessment protocols and data acquisition procedures, including variations in MRI scanner specifications and imaging parameters. Consequently, the ABIDE dataset inherently exhibits a non-independent and non-identically distributed (non-IID) data structure. Prior methodologies for processing ABIDE data have employed two primary strategies: 1) selecting subsets of large medical sites as FL clients \cite{li2020multi,dinsdale2022fedharmony,zhang2024preserving}, or 2) centralized aggregation followed by random data partitioning across clients \cite{peng2022fedni}. The first strategy neglects smaller contributing sites, whereas the second approach fails to preserve institutional heterogeneity due to potential inter-client data overlap from shared medical centers. To address these limitations, this paper proposes a geographically stratified non-IID partitioning strategy. The 17 sites are stratified into four geographically coherent regions: Europe, Western USA, Central USA, and Eastern USA (see Table \ref{tab:data_distribution}). This partitioning ensures that each FL client represents a unique geographic region, preserving data heterogeneity and minimizing inter-client overlap. The resulting dataset is referred to as FedASD.

\input{table/tab1}

\subsubsection{Skin Lesion Classification}
\textbf{Task Description}: Skin cancer, particularly melanoma, remains one of the most lethal malignancies. Early detection of melanoma not only can significantly reduce treatment costs but can enhance patient survival \cite{8333693,10.1007/978-3-030-87199-4_32}. Dermoscopy is one of the most widely adopted non-invasive imaging techniques to capture detailed morphological and visual characteristics of pigmented lesions for this purpose \cite{8333693}. In this study, we utilize dermoscopic images to diagnose skin lesions, framing the problem as a multi-class classification task.

\textbf{Dataset Description}: We utilized the publicly available Derm7pt dataset \cite{8333693}, which comprises dermoscopic images, clinical images, and associated meta-data. Dermoscopic images are obtained using a dermatoscope with a standardized field of view and controlled acquisition conditions. The patient meta-data encompasses additional details, such as gender and lesion location \cite{8333693}. In Derm7pt, 20 distinct skin lesion conditions were categorized into five primary types: basal cell carcinoma (BCC), nevus (NEV), melanoma (MEL), miscellaneous (MISC), and seborrheic keratosis (SK). Lesions within each category share similar clinical interpretations. The Derm7pt dataset consists of 1,011 cases, with 42 BCC, 575 NEV, 252 MEL, 97 MISC, and 45 SK lesions. All images have been resized to \(512 \times 512 \times 3\) pixels. Our focus is on diagnosing the five types of skin lesions using dermoscopic images.

\begin{figure}[hb]
\begin{minipage}[b]{1.0\linewidth}
    \centering
    \centerline{\includegraphics[width=\textwidth]{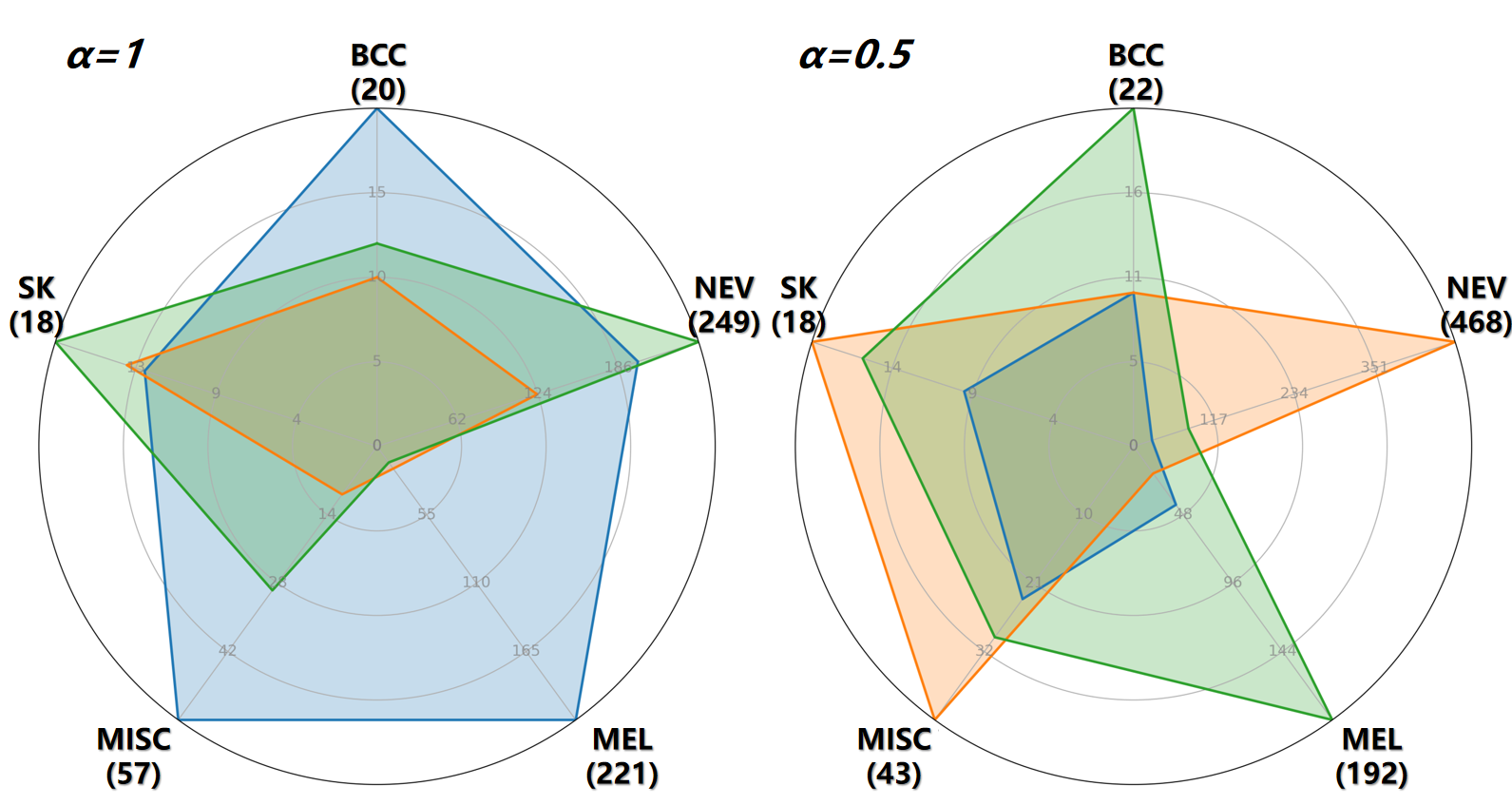}}

 \caption{\small{Distribution of non-IID data across the clients in our FedSkin datasets for skin lesion classification. Different colors represent different clients with unique non-IID distributions.}}
	\label{fig:fig4}
	\vspace{-15px}
 \end{minipage}
 \vspace{-10px}
\end{figure}

\textbf{Non-IID Dataset Construction}: The Skin dataset adheres to strict consistency in imaging acquisition protocols, with standardized scanner configurations and processing pipelines across all samples. To systematically model non-IID data distributions, we adopt a Dirichlet distribution ($Dir(\alpha)$), a probabilistic framework that enables controlled simulation of data heterogeneity by perturbing the original IID structure. Specifically, for a dataset with $C$ classes and $N$ clients, we generate an $N$-dimensional probability vector $p_c=\left \{ p_{c,1}, ..., p_{c,N}\right \} \sim Dir(\alpha)$ for each class $c$, where $\sum_{i=1}^N p_{c,i} = 1$ and each element $p_{c,i}$ represents the proportion of class $c$ samples assigned to client $i$. By partitioning class instances across FL clients based on the concentration parameter $\alpha$, this approach allows for precise control over data distribution skewness. Lower values of $\alpha$ induce a more pronounced skew, thereby accentuating non-IID characteristics. In our study, we construct three FL clients and implement two distinct non-IID scenarios using \( \alpha \in \{1.0, 0.5\}\), as shown in Figure \ref{fig:fig4}. This dual mechanism, encompassing both label and quantity skew, reflects real-world medical imaging constraints where class imbalances and uneven data collection practices are prevalent. The resulting dataset is designated as FedSkin.

\subsection{Experimental Setup} \label{sec:setup}
\subsubsection{Baselines}
We evaluate our method by conducting comprehensive comparisons against baselines from both model-homogeneous and model-heterogeneous FL scenarios, which include seven baselines. In the model-homogeneous FL setting, all clients are equipped with a ResNet-10 model featuring a fixed channel size of 64. In contrast, the model-heterogeneous FL setting employs the same ResNet-10 architecture across all clients but introduces variability in the initial channel sizes to simulate heterogeneity. Specifically, the first client is assigned a channel size of 64, the second client 62, and each subsequent client’s channel size decreases by 2, thereby progressively increasing the degree of model heterogeneity. Baselines from both model-homogeneous and model-heterogeneous FL are summarized in the following:

\noindent\textbf{Centralized}: An ideal scenario where all clients' data can be freely shared among clients without privacy constraints. Each client’s model is trained on the entire dataset assembled from all clients, serving as a theoretical upper-bound performance benchmark.

\noindent\textbf{Local}: In this setting, each client's model is trained solely on its own private dataset, with no communication between clients. This approach more accurately reflects real-world medical applications, where data privacy and security are paramount considerations.

\noindent\textbf{FedAvg}: A foundational FL algorithm where clients train models locally and aggregate parameters through model averaging after each communication round. FedAvg is designed for model-homogeneous settings and often struggles with non-IID data distributions.

\noindent\textbf{FedProx}: An extension of FedAvg that introduces a proximal term to regulate local model updates, enhancing convergence stability in heterogeneous data or system environments. However, it remains infeasible to model-homogeneous settings.

\noindent\textbf{FedBN}: This method addresses feature shifts in non-IID data by preserving client-specific batch normalization (BN) layers while aggregating other parameters globally. While effective in heterogeneous data environments, FedBN is constrained to model-homogeneous architectures.

\noindent\textbf{FedMRL}:A MHFL framework where a shared global auxiliary model collaborates with heterogeneous local models. Their representations are adaptively fused through a lightweight projector and further organized into multi-granularity Matryoshka representations for joint learning.

\noindent\textbf{pFedAFM}: A model-heterogeneous personalized FL framework that equips each client with a shared global auxiliary feature extractor alongside its local heterogeneous model. An iterative training strategy enables effective global–local knowledge exchange, while a trainable weight vector adaptively mixes features from both extractors to handle batch-level data heterogeneity.

\noindent\textbf{AlFeCo}: This framework enables knowledge exchange across clients without relying on public data. It facilitates collaborative updates of heterogeneous models on the server through a two-stage training process: first, an allosteric feature generator distills task-relevant knowledge into shared and client-specific codes; second, a dual-path (model–model and model–prediction) communication mechanism supervises model updates using exchanged codes and generated features, ensuring effective knowledge integration.

\noindent\textbf{FedCross}: This approach is considered the SOTA P2P FL approach that eliminates model aggregation by circulating a single global model among clients for sequential training. FedCross is designed for model-homogeneous settings and does not support architectural heterogeneity.

\noindent\textbf{FedCross$^{\dagger}$}: While FedCross was developed for homogeneous-model FL setting, its P2P characteristics allow its natural extension to heterogeneous-model FL setting. Thus, we adapted a variant of FedCross here (i.e. FedCross$^{\dagger}$) where we allow each client to maintain a personalized, architecturally heterogeneous model during P2P cross-client training. 

\subsubsection{Evaluation Strategies}
We use the area under the ROC curve (AUC) as the primary evaluation metric to assess classification performance, as AUC provides a more robust measure across varying thresholds and offers greater interpretability.

Given the limited sample sizes of the ABIDE and DERM7pt datasets, we adopt 5-fold cross-validation to report the final results. Specifically, for local performance, we report the average AUC across five folds for each client. For global performance, we first compute the average AUC across all centers within each fold, then report the mean and standard deviation across the five folds. Thus, we present both the local performance of each client and the overall global performance.

Additionally, we define two evaluation settings: Local Test and Global Test. In the Local Test, each client’s model is evaluated solely on its own local test dataset to assess its specialization capability. In the Global Test, each client’s model is evaluated on the test datasets of all clients, and the final performance is obtained by averaging the results across all clients, thereby assessing the model’s generalization ability. Please note that when computing the final performance in the Global Test, all clients’ test datasets are assigned equal weighting, regardless of their sample sizes, to mitigate potential arising from imbalanced data distributions across clients. Furthermore, it is worth noting that for several model-homogeneous methods—namely \text{Centralized}, \text{FedAvg}, \text{FedProx}, and \text{FedCross}—only a single shared model is maintained. Consequently, in the Global Test scenario, these methods yield only a mean performance result derived from 5-fold cross-validation, rather than individual global results for each client-specific model.

\subsubsection{Implementation Details}
All methods were trained to converge using a batch size of 8. For the ASD classification task, we configured the training process with 40 communication rounds, while for the skin lesion classification task, we used 45 communication rounds. Each round consisted of 5 training iterations. Notably, in P2P training scenarios, the number of training iterations per round was scaled to 5×N, where N denotes the number of clients, to account for the distributed nature of the learning process.

In the ASD classification task, the spatial dimensions of the four 3D temporal statistics were downscaled to $48 \times 48 \times 48$ and subsequently concatenated along the channel dimension to construct the input representation. For the skin lesion classification task, we exclusively employed dermoscopic images, which were resized to a uniform resolution of $224 \times 224$ pixels to ensure consistency across the dataset.

To enhance generalization and robustness, we applied standard spatial data augmentation techniques, e.g. random flipping, rotation, translation, and scaling, to both the 3D fMRI statistics and the 2D dermoscopy images. Optimization was performed using the Adam optimizer with a learning rate of $1 \times 10^{-4}$, aimed at minimizing the joint mutual learning loss as formulated in Equation \ref{eq:eq2}. Additionally, the balancing hyperparameter \(\gamma\) was empirically set to 1 for the ASD classification task and 100 for the skin lesion classification task.

\subsection{Experimental Results} \label{sec:results}
\input{table/tab2}
\input{table/tab3}

\subsubsection{Performance Comparison with Baselines}
Table \ref{tab:tb2} and Table \ref{tab:tb3} present a comprehensive evaluation of our proposed FedSKD against state-of-the-art baselines in both model-homogeneous and model-heterogeneous settings. The evaluation is conducted under Local Test and Global Test scenarios for our two tasks.

As shown in Table \ref{tab:tb2}, in model-heterogeneous setting, FedSKD achieves state-of-the-art global average performance for ASD Classification across both test settings, with mean AUC values of 76.66\% (Local Test) and 67.10\% (Global Test), outperforming FedCross$^{\dagger}$, the leading heterogeneous P2P FL baseline, by significant margins of 5.39\% and 3.44\%, respectively. Notably, FedSKD also surpasses recently proposed MHFL baselines, outperforming pFedAFM by 6.02\% and 4.00\%, FedMRL by 8.38\% and 6.30\%, and AlFeCo by 6.15\% and 6.05\%, in the Local and Global Tests, respectively. Furthermore, FedSKD exceeds our lower-bound baseline (Local) by substantial margins of 7.97\% and 6.05\% in terms of the Local and Global Tests, respectively, demonstrating its robust adaptation to heterogeneous data distributions. Similar superiority is observed in model-homogeneous settings, where FedSKD exceeds homogeneous FedCross by 5.39\% and 3.44\% while also significantly outperforming all homogeneous baselines.

The quantitative results summarized in Table \ref{tab:tb3} further highlight the superiority of FedSKD in Skin Lesion Classification under varying non-IID data distributions ($\alpha$=1 and $\alpha$=0.5). Notably, in model-heterogeneous setting, FedSKD consistently achieves the highest mean AUC values in both test scenarios, outperforming FedCross$^{\dagger}$ by 3.14\% and 3.05\% ($\alpha$=1) and by 3.43\% and 3.62\% ($\alpha$=0.5) for Local and Global Tests, respectively. Compared with the three most recent MHFL baselines, pFedAFM, FedMRL and AlFeCo, FedSKD achieves improvements of up to 8.05\% and 6.11\% ($\alpha$=1) and 8.62\% and 8.26\% ($\alpha$=0.5) over the strongest competitor, pFedAFM. The performance gaps widen further when compared to the Local baseline, with improvements of 8.14\% and 8.56\% ($\alpha$=1) and 9.86\% and 9.61\% ($\alpha$=0.5). Under the homogeneous configurations, FedSKD maintains substantial leads over FedCross (2.98\% and 2.93\% ($\alpha$=1) and by 4.44\% and 4.03\% ($\alpha$=0.5)) and other baselines.

Three critical insights emerge from these findings: First, aggregation-free P2P methods (FedSKD, FedCross variants) consistently outperform aggregation-based approaches (FedMRL, pFedAFM, AlFeCo, FedAvg, FedProx, FedBN) across all settings, highlighting the inherent limitations of parameter averaging in handling data heterogeneity. Second, FedSKD consistently outperforms FedCross in both homogeneous and heterogeneous settings, validating that our similarity knowledge distillation (SKD) mechanism effectively mitigates model drift and knowledge dilution through bidirectional feature-space regularization (i.e. multi-dimensional SKD). Third, the simultaneous improvement in Local Test (personalization) and Global Test (generalization) metrics demonstrates FedSKD's unique capability to balance client-specific adaptation with cross-client knowledge fusion.

\begin{figure}[htb!]
    \centering
    \subfloat[]{%
        \includegraphics[width=0.48\textwidth]{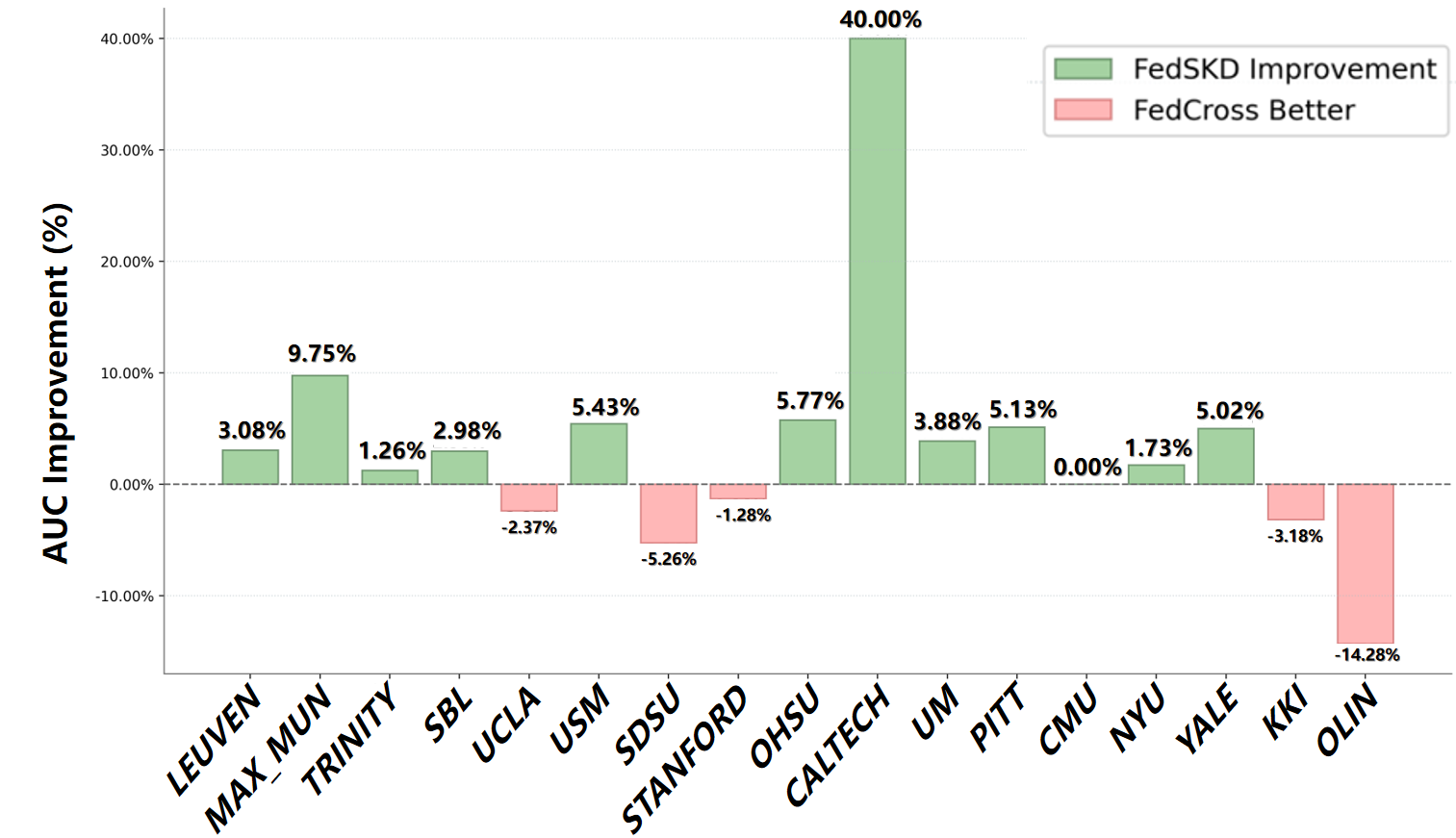}
    }
    \hfill
    \subfloat[]{%
        \includegraphics[width=0.45\textwidth]{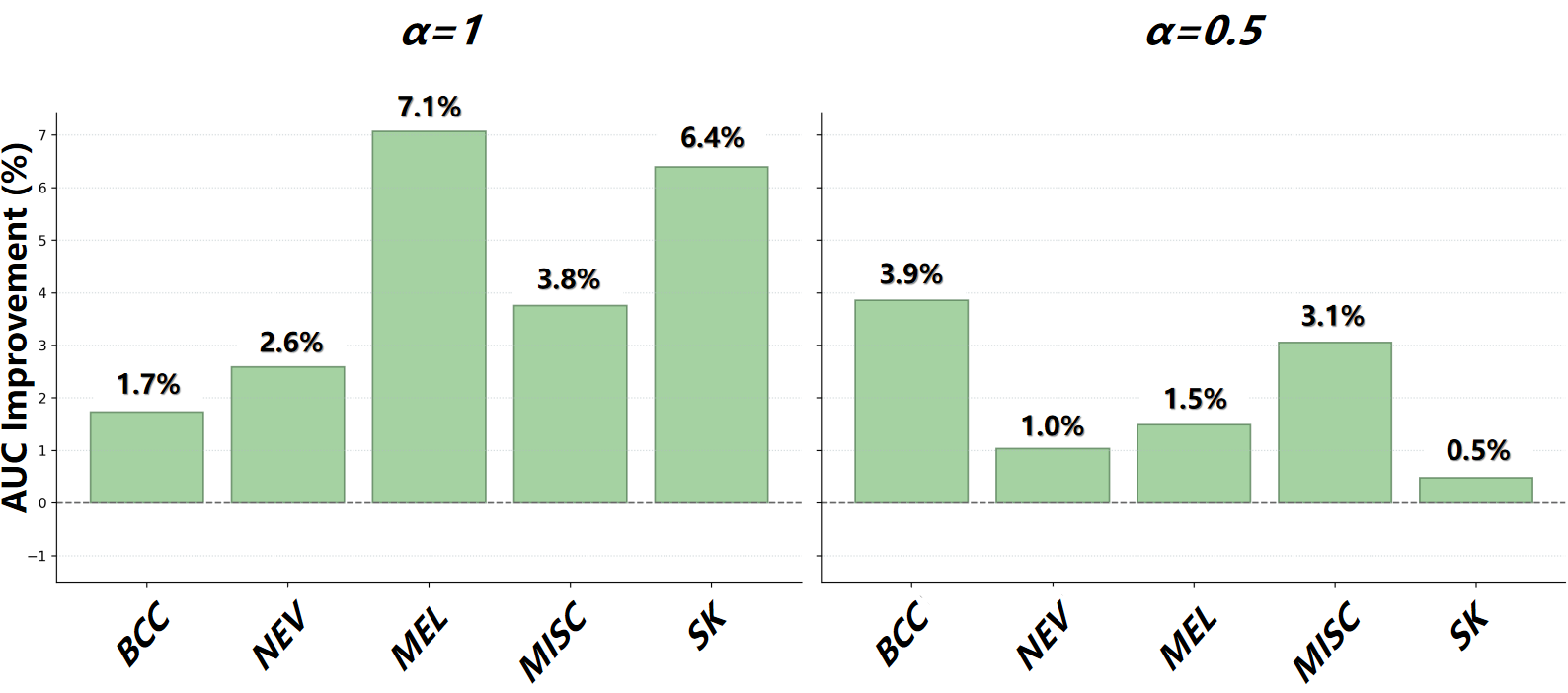}
    }
    \caption{Comparison of AUC Improvements: FedSKD vs. FedCross. (a) Summary of AUC performance gain across 17 institutions from the FedASD dataset, comparing FedSDK and FedCross. (b) Summary of AUC performance gain across 5 skin lesion types from the FedSkin dataset, comparing FedSDK and FedCross.}
    \label{fig:fig5}
\end{figure}

Consolidated predictions from all client models under 5-fold cross-validation reveal FedSKD's consistent superiority. For ASD classification across 17 institutions in the ABIDE dataset (Figure \ref{fig:fig5} (a)), FedSKD achieves higher accuracy in the majority of institutions compared to FedCross. In skin lesion diagnosis (Figure \ref{fig:fig5} (b)), FedSKD demonstrates uniform improvements across all five lesion types under both Non-IID settings ($\alpha$=1 and $\alpha$=0.5), with particularly significant gains in rare lesion categories (e.g. BCC and SK), underscoring its effectiveness in handling class imbalance.

These results collectively establish FedSKD as a paradigm-shifting framework that addresses fundamental limitations in both traditional P2P (such as model drift and knowledge dilution) and aggregation-based FL approaches (which suffer from performance degradation).

\subsubsection{Computational Overhead and Communication Cost}
On the FedASD dataset, we analyze the computational and communication costs of FedSKD in comparison with other baseline methods, as presented in Table \ref{tab:tb8}. For simplicity, each client adopts the same ResNet10 architecture with an initial channel size of 64. We report the GFLOPs for client-side training, the model size during training, the number of parameters exchanged per client during communication, and the parameters retained locally for deployment. The training GFLOPs are computed on a single fMRI sample of size $(4\times48\times48\times48)$, including both forward and backward passes (approximately twice the cost of forward propagation). 

\noindent\textbf{Computational Overhead:}  
FedSKD introduces higher training GFLOPs, primarily because it requires training two models locally, effectively doubling the computational overhead compared to FedAvg. However, the additional cost of our multi-dimensional similarity knowledge distillation is negligible. More importantly, compared to FedAvg and FedCross, FedSKD achieves consistently superior performance across both datasets (FedASD and FedSkin) and under both model configurations, making the extra computational overhead well justified. Although both pFedAFM and FedMRL also require training two models locally, pFedAFM’s two-stage training procedure reduces training efficiency, and neither method matches the performance gains of FedSKD. Notably, AlFeCo alternates between training feature generators and heterogeneous models on the server side, leading to substantial additional computational overhead. Furthermore, unlike aggregation-based methods, FedSKD is aggregation-free and imposes no additional computational load on the server side.

\noindent\textbf{Communication Cost:}  
FedSKD requires the same communication bandwidth and deployment model size as FedAvg, while incurring twice the local training parameters since two models are trained simultaneously. Nevertheless, the performance improvement over FedAvg is substantial, making the communication cost entirely acceptable. In contrast, pFedAFM and FedMRL not only train both a local heterogeneous model and a global homogeneous small model, but also retain both for deployment, thereby incurring extra storage and communication costs. By comparison, FedSKD strikes a more favorable balance, achieving significantly better performance without introducing additional deployment overhead.

\input{table/tab8}

\subsubsection{Ablation Studies}
\noindent\textbf{Component-wise Analysis of SKD Effectiveness:} Table \ref{tab:tb4} validates the contribution of each similarity dimension in our multi-dimensional SKD through model-heterogeneous ablation experiments. The results demonstrate that the complete SKD implementation, which integrates all similarity learning, achieves optimal performance across both tasks, highlighting the complementary nature of multi-granularity feature alignment. Notably, batch-wise SKD provides the most significant performance boost compared to pixel/voxel-wise components, as inter-sample correlations prove more discriminative than fine-grained spatial correspondences for image-level classification tasks. Furthermore, even variants with partial SKD configurations consistently outperform FedCross (in Table \ref{tab:tb2}-\ref{tab:tb3}), particularly in Local Test scenarios, underscoring the robustness of our method against knowledge fragmentation in peer-to-peer FL. The progressive performance improvement from single-to-full SKD configurations demonstrates FedSKD's ability to synergistically address model drift through hierarchical feature alignment.

\input{table/tab4}

\noindent\textbf{Layer-wise Analysis of SKD Effectiveness:} To investigate how hierarchical feature distillation impacts the DAM-KTM mutual learning within our FedSKD framework, we conduct a layer-wise ablation study under model-heterogeneous setting. Building on the established principle that deep networks progressively encode features from low-level textures to high-level semantics \cite{zeiler2014visualizing}, we hypothesize that deeper layers capture more task-aligned semantic representations. Therefore, our experimental design initiates from the deepest layer (Layer 4 of ResNet-10), where feature embeddings exhibit the strongest correlation with classification objectives, and incrementally incorporates shallower layers (Layer 3 $\rightarrow$ Layer 2 $\rightarrow$ Layer 1) to systematically evaluate multi-scale feature alignment.

\input{table/tab5}

As shown in Table \ref{tab:tb5}, performance metrics for both ASD and skin lesion classification improve progressively as SKD engages broader network depths. The full configuration—utilizing all four ResNet-10 layers—achieves peak performance, demonstrating that knowledge transfer across clients is significantly enhanced when multi-level feature representations participate in the distillation process. Specifically, while deep layers encode task-specific semantics critical for diagnostic decision-making, shallow layers preserve spatial and textural biomarkers essential for robust generalization. This hierarchical synergy underscores the importance of multi-level feature alignment in federated learning, where both localized and global representations contribute to mitigating model drift and improving cross-client knowledge fusion.

\vspace{1mm}
\noindent\textbf{Timing for Multi-dimensional SKD Integration:} To determine the optimal timing for introducing our multi-dimensional SKD into the FedSKD framework, we conduct a study varying the timing to activate multi-dimensional SKD, including 0\% (training onset), 25\%, 50\%, and 75\% of total communication rounds. Please note here that prior to SKD activation, models follow standard P2P transfer with target client adaptation. 
As Table \ref{tab:tb6} demonstrates, a strong temporal dependency emerges: earlier SKD integration yields progressively greater performance gains, with introducing SKD at training onset achieves peak performance across both tasks and across both evaluation settings. By constraining parameter divergence through early feature-space (i.e. multi-dimensional SKD) regularization, FedSKD prevents irreversible model drift while preserving client-specific knowledge. The performance degradation observed with delayed SKD activation underscores the importance of proactive knowledge preservation in P2P FL systems.

\input{table/tab9}
\noindent\textbf{Robustness Under Data Poisoning Attacks:} Federated learning systems are vulnerable to poisoning attacks, where malicious clients manipulate data or models to compromise collaborative training. In this section, we examine data poisoning in the form of a label flipping attack, in which an adversarial client intentionally corrupts ground-truth labels. Under the model heterogeneous FedASD setting, we designate the third client (Middle USA) as the attacker and flip all labels in its local dataset, converting ASD to Normal and Normal to ASD. We then evaluate all MHFL methods under this extreme setting to assess how the malicious client affects the performance of all participating models.

As shown in Table~\ref{tab:tb9}, FedSKD remains substantially more resilient than all competing MHFL methods under the label-flipping attack and even surpasses the performance achieved by some baselines in the absence of attacks. Notably, the malicious client ($M_{3}$) under FedSKD still maintains a high AUC on its unpoisoned local test set, whereas other methods exhibit severe degradation due to the corrupted labels. This highlights the robustness of our peer-to-peer framework and the proposed multi-dimensional similarity knowledge distillation mechanism in mitigating the impact of data poisoning attacks.

\input{table/tab6}

\input{table/tab7}

\noindent\textbf{Impact on Fairness Improvement:} We further investigate the effect of FedSDK on improving fairness, and we focus on outcome-consistency group fairness \cite{xu2023fairadabn,xu2024apple,jin2024fairmedfm,xu2024addressing} with respect to the sex attribute (male vs. female) here. For this analysis, we consider only the local test setting, where each client's model is evaluated exclusively on its own dataset to measure fairness. We employ the maximum AUC gap ($\Delta_{auc}$) as our fairness metric, defined as the difference between the highest and lowest AUC values across sex subgroups in Equation \ref{eq:eq16}, to measure disparities in model performance:

\begin{equation} 
\Delta_{\text{auc}} = \left| \frac{1}{N} \sum_{k=1}^{N} \text{AUC}_{\text{male}}^k - \frac{1}{N} \sum_{k=1}^{N} \text{AUC}_{\text{female}}^k \right|
\label{eq:eq16}
\end{equation}
where $\text{AUC}_{\text{male}}^k$ and $\text{AUC}_{\text{female}}^k$ represent the AUC values for male and female subgroups, respectively, computed on the dataset $D_k$ of the $k$-th client; $N$ is the total number of clients; and $|\cdot|$ denotes the absolute value operation. We compare the sex fairness of our FedSKD with FedCross, on both classification tasks. As we can see from Table \ref{tab:tb7}, the results demonstrate that our method not only improves the average AUC but also significantly reduces the disparity between them. For instance, in the ASD classification task, our approach achieves a 6.37\% improvement in fairness, effectively mitigating unfairness related to the sensitive attribute. 

%% file: table/tab1.tex

\begin{table}[h]
\caption{Distribution of non-IID data across clients in our FedASD dataset. Four geographically different regions are clustered and defined as four different clients.}
\centering
\resizebox{\columnwidth}{!}{
\begin{tabular}{cllcc}
\toprule
\toprule
\textbf{Client} & \textbf{Site} & \textbf{Region} & \textbf{Diagnosis (ASD/TC)} & \textbf{Total (ASD/TC)} \\ \midrule
\multirow{4}{*}{1 (Europe)} & LEUVEN   & Belgium     & 26/30              & \multirow{4}{*}{172 (76/96)} \\ 
                            & MAX\_MUN & Germany     & 19/27              &                           \\ 
                            & TRINITY  & Ireland     & 19/25              &                           \\ 
                            & SBL      & Netherlands & 12/14              &                           \\ \midrule
\multirow{6}{*}{2 (West USA)} & UCLA     & CA, USA     & 48/37              & \multirow{6}{*}{244 (128/116)} \\ 
                            & USM      & UT, USA     & 43/24              &                           \\ 
                            & SDSU     & CA, USA     & 8/19               &                           \\ 
                            & STANFORD & CA, USA     & 12/13              &                           \\ 
                            & OHSU     & OR, USA     & 12/13              &                           \\ 
                            & CALTECH  & CA, USA     & 5/10               &                           \\ \midrule
\multirow{3}{*}{3 (Middle USA)} & UM       & MI, USA     & 47/73              & \multirow{3}{*}{181 (77/104)} \\ 
                                & PITT     & PA, USA     & 24/26              &                           \\ 
                                & CMU      & PA, USA     & 6/5                &                           \\ \midrule
\multirow{4}{*}{4 (East USA)}   & NYU      & NY, USA     & 74/98              & \multirow{4}{*}{274 (122/152)} \\ 
                                & YALE     & CT, USA     & 22/19              &                           \\ 
                                & KKI      & MD, USA     & 12/21              &                           \\ 
                                & OLIN     & CT, USA     & 14/14              &                           \\ \bottomrule\bottomrule
\end{tabular}
}
\label{tab:data_distribution}
\end{table}


%% file: table/tab2.tex
\begin{table*}[htb!]
\caption{Performance comparison across methods in the local and global test scenarios on the FedASD dataset. The best results under the model-heterogeneous setting are highlighted in bold. The symbol $^{*}$ indicates that the performance difference between our FedSKD and the second-best baseline methods under the same FL setting is statistically significant at $p<0.05$.}
\centering
\resizebox{\textwidth}{!}{
\begin{tabular}{c|ccccc|ccccc}
\hline
\hline
Method & \multicolumn{5}{c|}{Local Test} & \multicolumn{5}{c}{Global Test} \\ \cline{2-11} 
(FedASD) & $M_{Europe}$ & $M_{WestUS}$ & $M_{CentralUS}$ & $M_{EastUS}$ & Mean & $M_{Europe}$ & $M_{WestUS}$ & $M_{CentralUS}$ & $M_{EastUS}$ & Mean \\ \hline
\multicolumn{1}{c|}{} & \multicolumn{10}{c}{\textbf{Homogeneous Models}} \\ \hline
Centralized & 70.27$\pm$5.40  & 71.02$\pm$6.14  & 75.55$\pm$5.43  & 72.92$\pm$5.40  & 72.44$\pm$2.27  & \textbackslash  & \textbackslash  & \textbackslash  & \textbackslash  & 67.65$\pm$2.30 \\ \hline
Local  & 66.40$\pm$6.48  & 68.06$\pm$3.93  & 66.41$\pm$4.15  & 70.23$\pm$4.64  & 67.77$\pm$2.69  & 59.53$\pm$4.00  & 62.60$\pm$2.78  & 62.10$\pm$2.85  & 60.47$\pm$2.74  & 61.17$\pm$2.05 \\ \hline

FedAvg & 69.21$\pm$6.21  & 73.18$\pm$2.05  & 67.94$\pm$6.61  & 69.36$\pm$4.86  & 69.92$\pm$2.55  & \textbackslash  & \textbackslash  & \textbackslash  & \textbackslash  & 64.70$\pm$1.79 \\ \hline
FedProx & 65.62$\pm$5.31  & 70.36$\pm$4.13  & 67.06$\pm$4.67  & 69.01$\pm$5.08  & 68.01$\pm$1.61  & \textbackslash  & \textbackslash  & \textbackslash  & \textbackslash  & 64.55$\pm$3.16 \\ \hline
FedBN & 69.48$\pm$4.83  & 68.69$\pm$3.53  & 72.47$\pm$3.89  & 69.52$\pm$8.01  & 70.04$\pm$2.88  & 65.48$\pm$1.99  & 66.15$\pm$1.94  & 65.57$\pm$2.55  & 64.31$\pm$2.97  & 65.38$\pm$2.18 \\ \hline

FedMRL  & 66.88$\pm$4.01  & 68.09$\pm$6.33  & 67.79$\pm$8.91  & 68.72$\pm$5.04  & 67.87$\pm$4.15  & 58.02$\pm$1.29  & 59.63$\pm$4.82  & 60.82$\pm$1.49  & 61.16$\pm$2.68  & 59.91$\pm$1.68 \\ \hline

pFedAFM  & 71.55$\pm$10.91  & 71.43$\pm$4.13  & 69.70$\pm$7.98  & 75.50$\pm$4.12  & 72.05$\pm$3.75  & 61.07$\pm$4.78  & 64.32$\pm$2.99  & 62.93$\pm$3.11  & 64.52$\pm$3.89  & 63.21$\pm$2.90 \\ \hline

AlFeCo  & 72.39$\pm$3.44  & 71.20$\pm$3.08  & 74.22$\pm$3.50  & 69.98$\pm$4.20  & 71.95$\pm$1.21  & 59.14$\pm$2.31  & 60.00$\pm$2.12  & 61.84$\pm$2.65  & 61.23$\pm$3.96  & 60.55$\pm$1.18 \\ \hline

FedCross & 72.75$\pm$5.01 & 72.23$\pm$6.96 & 78.00$\pm$3.77 & 69.92$\pm$5.87 & 73.23$\pm$2.70 & \textbackslash & \textbackslash & \textbackslash & \textbackslash & 66.12$\pm$2.97 \\ \hline

\rowcolor{gray!20} FedSKD & 79.80$\pm$7.63  & 76.30$\pm$7.16  & 81.20$\pm$3.19  & 73.93$\pm$4.66  & 77.81$\pm$1.72  & 66.49$\pm$2.66  & 67.83$\pm$3.24  & 68.18$\pm$3.10  & 67.70$\pm$3.02  & 67.55$\pm$2.75 \\ \hline
\multicolumn{1}{c|}{} & \multicolumn{10}{c}{\textbf{Heterogeneous Models}} \\ \hline

Centralized & 66.76$\pm$7.35  & 72.87$\pm$3.91  & 75.05$\pm$2.19  & 69.18$\pm$6.44  & 70.97$\pm$2.12  & 65.85$\pm$2.54  & 67.13$\pm$1.90  & 65.18$\pm$1.68  & 65.91$\pm$2.20  & 66.02$\pm$1.79 \\ \hline
Local  & 68.90$\pm$2.13  & 67.06$\pm$4.16  & 69.74$\pm$5.30  & 69.03$\pm$4.07  & 68.69$\pm$2.73  & 60.34$\pm$1.68  & 61.01$\pm$2.85  & 61.78$\pm$1.07  & 61.08$\pm$1.56  & 61.05$\pm$1.23 \\ \hline

FedMRL  & 67.15$\pm$5.47  & 69.34$\pm$5.52  & 66.85$\pm$6.18  & 69.78$\pm$3.96  & 68.28$\pm$3.64  & 59.26$\pm$2.11  & 62.67$\pm$3.05  & 61.16$\pm$3.41  & 60.11$\pm$3.18  & 60.80$\pm$2.30 \\ \hline

pFedAFM  & 70.02$\pm$7.15  & 70.76$\pm$5.36  & 69.36$\pm$7.87  & 72.54$\pm$4.89  & 70.67$\pm$1.75  & 59.58$\pm$4.81  & 63.65$\pm$3.33  & 64.82$\pm$2.83  & 64.35$\pm$3.32  & 63.10$\pm$2.40 \\ \hline

AlFeCo  & 71.73$\pm$4.24  & 67.29$\pm$6.15  & 73.25$\pm$3.61  & 69.79$\pm$3.81  & 70.51$\pm$1.49  & 59.96$\pm$3.24  & 61.49$\pm$1.02  & 60.53$\pm$2.72  & 61.05$\pm$3.32  & 61.05$\pm$2.01 \\ \hline

FedCross$^{\dagger}$ & 70.67$\pm$3.11  & 70.13$\pm$5.69  & 72.67$\pm$3.25  & 71.60$\pm$5.78  & 71.27$\pm$3.17  & 63.33$\pm$4.33  & 62.89$\pm$5.16  & 63.90$\pm$2.68  & 63.51$\pm$3.54  & 63.41$\pm$3.57 \\ \hline

\rowcolor{gray!20} FedSKD & 75.83$\pm$4.22  & 76.14$\pm$4.30  & 79.83$\pm$6.68  & 74.86$\pm$4.45  & \textbf{76.66$\pm$2.70}$^{*}$  & 67.44$\pm$2.13  & 66.36$\pm$2.29  & 67.93$\pm$3.02  & 66.69$\pm$3.47  & \textbf{67.10$\pm$2.23}$^{*}$ \\ \hline \hline
\end{tabular}
}
\label{tab:tb2}
\end{table*}

%% file: table/tab3.tex
\begin{table*}[htb!]
        \caption{Performance comparison across methods in the local and global test scenarios on the FedSkin datasets. The best results under the model-heterogeneous setting are highlighted in bold. The symbol $^{*}$ indicates that the performance difference between our FedSKD and the second-best baseline methods under the same FL setting is statistically significant at $p<0.02$.}
        \centering
        \tiny
        \resizebox{\textwidth}{!}{
        \begin{tabular}{c|cccc|cccc}
        \hline
        \hline
        Method & \multicolumn{4}{c|}{Local Test} & \multicolumn{4}{c}{Global Test} \\ \cline{2-9} 
        FedSkin ($\alpha$=1) & \multicolumn{1}{c}{$M_1$}        & \multicolumn{1}{c}{$M_2$}        & \multicolumn{1}{c}{$M_3$}        & Mean            & \multicolumn{1}{c}{$M_1$}        & \multicolumn{1}{c}{$M_2$}        & \multicolumn{1}{c}{$M_3$}        & Mean            \\ \hline
        \multicolumn{1}{c|}{} & \multicolumn{8}{c}{\textbf{Homogeneous Models}} \\ \hline                                                                   
        Centralized     & 86.41$\pm$3.31 & 85.67$\pm$4.63 & 85.40$\pm$5.61 & 85.83$\pm$1.68 & \textbackslash & \textbackslash & \textbackslash & 82.43$\pm$1.91       \\ \hline
        
        Local      & 81.06$\pm$6.48 & 81.97$\pm$9.21 & 76.76$\pm$7.04 & 79.93$\pm$4.18 & 79.61$\pm$3.44 & 75.84$\pm$2.32 & 72.13$\pm$3.66 & 75.86$\pm$2.47     \\ \hline
        
        FedAvg     & 84.52$\pm$2.73 & 82.30$\pm$7.47 & 84.97$\pm$6.06 & 83.93$\pm$3.39 & \textbackslash & \textbackslash & \textbackslash & 81.67$\pm$3.44     \\ \hline
        FedProx    & 82.97$\pm$3.91 & 79.83$\pm$6.91 & 83.86$\pm$4.81 & 82.22$\pm$3.63 & \textbackslash & \textbackslash & \textbackslash3 & 79.63$\pm$4.13     \\ \hline
        FedBN      & 81.47$\pm$6.55 & 78.68$\pm$4.66 & 81.42$\pm$5.15 & 80.52$\pm$2.66 & 78.60$\pm$3.57 & 75.83$\pm$5.29 & 76.29$\pm$3.82 & 76.91$\pm$3.52     \\ \hline

        FedMRL  & 76.71$\pm$4.30  & 78.57$\pm$9.67  & 77.04$\pm$8.01  & 77.44$\pm$5.27  & 77.92$\pm$3.05  & 75.09$\pm$2.63  & 71.53$\pm$3.18  & 74.85$\pm$2.55  \\ \hline

        pFedAFM  & 85.32$\pm$2.80  & 79.76$\pm$12.22  & 78.06$\pm$7.37  & 81.05$\pm$4.18  & 79.94$\pm$2.09  & 75.12$\pm$2.63  & 72.71$\pm$1.60  & 75.92$\pm$0.58   \\ \hline

        AlFeCo  & 81.78$\pm$3.28  & 78.34$\pm$5.22  & 80.84$\pm$7.28  & 80.32$\pm$3.66  & 80.41$\pm$3.67  & 73.18$\pm$2.67  & 75.01$\pm$2.30  & 76.20$\pm$0.22   \\ \hline
        
        FedCross   & 83.56$\pm$4.65 & 88.41$\pm$4.73 & 84.56$\pm$6.57 & 85.51$\pm$3.66 & \textbackslash & \textbackslash & \textbackslash & 81.66$\pm$2.56     \\ \hline

        \rowcolor{gray!20} FedSKD  & 87.28$\pm$3.70 & 91.36$\pm$4.67 & 86.82$\pm$5.90 & 88.49$\pm$1.35 & 84.35$\pm$0.84 & 84.33$\pm$1.83 & 85.10$\pm$0.70 & 84.59$\pm$0.87     \\ \hline \hline

        \multicolumn{1}{c|}{} & \multicolumn{8}{c}{\textbf{Heterogeneous Models}} \\ \hline 
        Centralized     & 86.37$\pm$3.82 & 86.15$\pm$5.28 & 86.18$\pm$5.40 & 86.23$\pm$1.94 & 83.32$\pm$1.31 & 82.61$\pm$1.75 & 83.79$\pm$2.53 & 83.24$\pm$1.36       \\ \hline
        
        Local      & 81.35$\pm$3.90 & 81.95$\pm$7.75 & 78.77$\pm$7.63 & 80.69$\pm$3.91 & 78.47$\pm$3.70 & 75.44$\pm$2.27 & 73.21$\pm$2.73 & 75.70$\pm$2.05     \\ \hline
        
        FedMRL  & 79.79$\pm$2.22  & 78.97$\pm$11.49  & 78.36$\pm$7.11  & 79.04$\pm$5.74  & 78.89$\pm$1.52  & 75.27$\pm$4.31  & 73.00$\pm$2.53  & 75.72$\pm$2.09  \\ \hline

        pFedAFM  & 83.72$\pm$2.89  & 80.51$\pm$13.05  & 78.11$\pm$7.66  & 80.78$\pm$5.04  & 81.03$\pm$2.74  & 78.65$\pm$3.24  & 74.77$\pm$2.99  & 78.15$\pm$2.49   \\ \hline

        AlFeCo  & 80.77$\pm$3.23  & 77.77$\pm$6.90  & 77.88$\pm$10.36  & 78.81$\pm$3.39  & 80.76$\pm$2.35  & 74.15$\pm$2.25  & 73.59$\pm$2.15  & 76.17$\pm$1.27   \\ \hline
        
        FedCross$^{\dagger}$   & 82.47$\pm$3.04 & 88.18$\pm$4.71 & 85.41$\pm$4.98 & 85.69$\pm$1.64 & 81.11$\pm$1.59 & 81.19$\pm$1.13 & 81.33$\pm$1.31 & 81.21$\pm$1.02   \\ \hline
        
        \rowcolor{gray!20} FedSKD  & 88.41$\pm$3.04 & 90.50$\pm$3.50 & 87.59$\pm$4.89 & \textbf{88.83$\pm$1.85}$^{*}$ & 84.56$\pm$1.69 & 84.58$\pm$2.42 & 83.65$\pm$2.66 & \textbf{84.26$\pm$2.13}$^{*}$     \\ \hline \hline

        \hline
        \hline
        Method & \multicolumn{4}{c|}{Local Test} & \multicolumn{4}{c}{Global Test} \\ \cline{2-9} 
        FedSkin ($\alpha$=0.5) & \multicolumn{1}{c}{$M_1$}        & \multicolumn{1}{c}{$M_2$}        & \multicolumn{1}{c}{$M_3$}        & Mean            & \multicolumn{1}{c}{$M_1$}        & \multicolumn{1}{c}{$M_2$}        & \multicolumn{1}{c}{$M_3$}        & Mean            \\ \hline
        \multicolumn{1}{c|}{} & \multicolumn{8}{c}{\textbf{Homogeneous Models}} \\ \hline                                                                   
        Centralized     & 86.40$\pm$3.25 & 82.42$\pm$3.47 & 86.14$\pm$4.73 & 84.99$\pm$2.36 & \textbackslash & \textbackslash & \textbackslash & 82.23$\pm$2.49     \\ \hline
        
        Local      & 80.93$\pm$4.67 & 75.38$\pm$5.20 & 78.84$\pm$2.96 & 78.38$\pm$1.98 & 75.25$\pm$2.85 & 70.54$\pm$1.15 & 78.25$\pm$2.14 & 74.68$\pm$1.48     \\ \hline
        
        FedAvg     & 82.03$\pm$4.80 & 79.87$\pm$2.05 & 81.46$\pm$3.14 & 81.12$\pm$1.42 & \textbackslash & \textbackslash & \textbackslash & 78.82$\pm$2.18     \\ \hline
        
        FedProx    & 83.08$\pm$4.22 & 78.23$\pm$6.22 & 78.48$\pm$2.66 & 79.93$\pm$1.99 & \textbackslash & \textbackslash & \textbackslash & 77.46$\pm$2.05     \\ \hline
        
        FedBN      & 79.06$\pm$6.27 & 78.23$\pm$3.99 & 76.01$\pm$3.33 & 77.77$\pm$2.49 & 75.35$\pm$3.59 & 75.99$\pm$3.51 & 75.31$\pm$3.70 & 75.55$\pm$3.23     \\ \hline

        FedMRL  & 75.83$\pm$8.86  & 78.23$\pm$3.22  & 80.88$\pm$2.89  & 78.31$\pm$3.76  & 71.56$\pm$5.36  & 72.85$\pm$2.21  & 77.32$\pm$2.01  & 73.91$\pm$1.91  \\ \hline

        pFedAFM  & 79.70$\pm$5.78  & 79.18$\pm$4.74  & 80.34$\pm$2.99  & 79.74$\pm$3.43  & 74.27$\pm$3.51  & 73.11$\pm$2.93  & 78.55$\pm$2.06  & 75.31$\pm$2.19   \\ \hline

        AlFeCo  & 76.14$\pm$6.31  & 80.22$\pm$3.05  & 79.88$\pm$2.72  & 78.75$\pm$3.40  & 72.86$\pm$4.35  & 74.32$\pm$1.74  & 77.53$\pm$1.89  & 74.90$\pm$2.42   \\ \hline
        
        FedCross   & 87.30$\pm$3.76 & 79.84$\pm$4.67 & 81.62$\pm$2.85 & 82.92$\pm$2.87 & \textbackslash & \textbackslash & \textbackslash & 79.49$\pm$3.54     \\ \hline

        \rowcolor{gray!20} FedSKD  & 89.15$\pm$3.62 & 86.65$\pm$4.64 & 86.29$\pm$3.09 & 87.36$\pm$2.79 & 83.70$\pm$1.67 & 83.10$\pm$2.20 & 83.75$\pm$1.89 & 83.52$\pm$1.80 \\ \hline \hline

        \multicolumn{1}{c|}{} & \multicolumn{8}{c}{\textbf{Heterogeneous Models}} \\ \hline    
        Centralized     & 88.14$\pm$1.74 & 82.21$\pm$5.77 & 85.74$\pm$4.39 & 85.37$\pm$2.23 & 82.68$\pm$2.30 & 82.16$\pm$3.39 & 80.93$\pm$2.49 & 81.93$\pm$2.54     \\ \hline
        
        Local      & 79.95$\pm$4.23 & 72.45$\pm$6.05 & 80.06$\pm$2.23 & 77.49$\pm$3.13 & 73.99$\pm$3.49 & 71.36$\pm$3.01 & 76.73$\pm$2.64 & 74.03$\pm$1.32     \\ \hline

        FedMRL  & 74.13$\pm$7.22  & 79.35$\pm$3.37  & 79.62$\pm$2.14  & 77.70$\pm$2.49  & 71.08$\pm$3.63  & 75.29$\pm$2.05  & 77.73$\pm$2.14  & 74.70$\pm$1.19  \\ \hline

        pFedAFM  & 76.19$\pm$6.81  & 78.01$\pm$6.73  & 82.00$\pm$2.87  & 78.73$\pm$2.95  & 72.26$\pm$3.88  & 74.31$\pm$4.49  & 79.57$\pm$2.16  & 75.38$\pm$1.36   \\ \hline

        AlFeCo  & 79.20$\pm$7.72  & 80.83$\pm$4.72  & 80.10$\pm$2.69  & 80.05$\pm$4.02  & 73.50$\pm$4.74  & 75.42$\pm$3.24  & 78.07$\pm$1.98  & 75.67$\pm$2.04   \\ \hline
        
        FedCross$^{\dagger}$   & 86.14$\pm$2.77 & 82.83$\pm$2.03 & 82.78$\pm$3.00 & 83.92$\pm$2.07 & 80.32$\pm$1.71 & 80.01$\pm$3.28 & 79.73$\pm$1.34 & 80.02$\pm$1.74 \\ \hline
        
        \rowcolor{gray!20} FedSKD  & 89.27$\pm$2.94 & 85.40$\pm$2.80 & 87.37$\pm$1.25 & \textbf{87.35$\pm$1.52}$^{*}$ & 84.27$\pm$1.78 & 82.61$\pm$2.18 & 84.05$\pm$1.87 & \textbf{83.64$\pm$1.81}$^{*}$ \\ \hline \hline

        \end{tabular}
        }
    
    \label{tab:tb3}
\end{table*}

%% file: table/tab8.tex
\begin{table}[htbp]
\centering
\caption{Computational and Communication Cost Comparison of FedSKD and Baseline Methods. Params are in M.}
\label{tab:fl_comparison_advanced}
\begin{tabular}{cccccc}
\toprule \toprule
\textbf{Method} & \textbf{\shortstack{Server\\Aggregation}} & \textbf{\shortstack{Train\\GFLOPs}} & \textbf{\shortstack{Train\\Params}} & \textbf{\shortstack{Commu\\Params}} & \textbf{\shortstack{Deploy\\Params}} \\
\midrule
FedAvg          & \textbf{\checkmark}              & 265.67                & 14.37                 & 28.73                 & 14.37                  \\
FedMRL          & \textbf{\checkmark}              & 485.69                & 26.67                 & 28.73                 & 26.67                  \\
pFedAFM         & \textbf{\checkmark}              & 110.99                & 26.17                 & 23.60                 & 26.17                  \\

AlFeCo & \textbf{\checkmark}              & 713.49                & 14.54                 & 28.73                 & 14.37                  \\

FedCross        & $\times$              & 265.67                & 14.37                 & 28.73                 & 14.37                  \\
FedSKD          &  $\times$             & 531.35                & 28.73                 & 28.73                 & 14.37                  \\
\bottomrule \bottomrule
\end{tabular}
\vspace{0.2cm}
\label{tab:tb8}
\end{table}

%% file: table/tab4.tex
\begin{table}[htb!]
    \caption{Ablation study on distillation types in multi-dimensional SKD. (a) Results on the FedASD dataset. (b) Results on the FedSkin datasets.}
    \centering
    \begin{minipage}[t]{\columnwidth}
        \centering
        \subfloat[]{
        \resizebox{0.8\columnwidth}{!}{
        \begin{tabular}{ccc|cc|cc}
        \toprule \toprule
        \multicolumn{3}{c|}{Similarity} & \multicolumn{2}{c|}{Local Test (\%)} & \multicolumn{2}{c}{Global Test (\%)} \\ \cmidrule{1-7} 
          Batch & Voxel & Region & Mean & Std & Mean & Std \\ \midrule
         \textbf{\checkmark} &  &  & 74.61 & 1.69 & 65.115 & 3.12 \\ 
         & \textbf{\checkmark} &  & 73.85 & 3.32 & 65.53 & 2.92 \\ 
         & & \textbf{\checkmark} & 74.71 & 2.40 & 66.97 & 2.27 \\
         \textbf{\checkmark} & \textbf{\checkmark} &  & 75.34 & 2.65 & 66.31 & 2.53 \\ 
         \textbf{\checkmark} &  & \textbf{\checkmark} & 75.45 & 1.92 & 67.00 & 2.50 \\ 
         \textbf{\checkmark} & \textbf{\checkmark} & \textbf{\checkmark} & 76.66 & 2.70 & 67.10 & 2.23 \\  \bottomrule \bottomrule
        \end{tabular}
        \label{tab:4a}
        }
        }
        
    \end{minipage}

    \begin{minipage}[t]{\columnwidth}
        \centering
        \subfloat[]{
        \tiny
        \resizebox{0.8\columnwidth}{!}{
        \begin{tabular}{cc|cc|cc}
        \toprule \toprule
        \multicolumn{2}{c|}{Similarity} & \multicolumn{2}{c|}{Local Test (\%)} & \multicolumn{2}{c}{Global Test (\%)} \\ \cmidrule{1-6}
        Batch & Pixel & Mean & Std & Mean & Std \\ \hline
        
        \multicolumn{6}{c}{$\alpha$=1} \\ \hline
        \textbf{\checkmark} & & 87.86 & 1.97 & 84.01 & 1.49 \\ 
        & \textbf{\checkmark}  & 87.29 & 2.04 & 82.62 & 1.61 \\ 
        \textbf{\checkmark} & \textbf{\checkmark} & 88.83 & 1.85 & 84.26 & 2.13 \\ 
        \hline
        \multicolumn{6}{c}{$\alpha$=0.5} \\ \hline
         \textbf{\checkmark} & & 86.06 & 1.50 & 82.60 & 1.36 \\ 
         & \textbf{\checkmark}  & 84.95 & 2.20 & 81.52 & 2.29 \\ 
         \textbf{\checkmark} & \textbf{\checkmark} & 87.35 & 1.52 & 83.64 & 1.81 \\  \bottomrule \bottomrule
        \end{tabular}
        \label{tab:4b}
        }
        } 
    \end{minipage}
    \label{tab:tb4}
\end{table}

%% file: table/tab5.tex
\begin{table}[htb!]
    \caption{Ablation study on activated layers in similarity knowledge distillation. (a) Results on the FedASD dataset. (b) Results on the FedSkin datasets.}
    \centering
    \begin{minipage}[t]{\columnwidth}
        \centering
        \subfloat[]{
        \resizebox{0.75\columnwidth}{!}{
        \begin{tabular}{cccc|cc|cc}
        \toprule \toprule
        \multicolumn{4}{c|}{Layer} & \multicolumn{2}{c|}{Local Test (\%)} & \multicolumn{2}{c}{Global Test (\%)} \\ \cmidrule{1-8} 
          1 & 2 & 3 & 4 & Mean & Std & Mean & Std \\ \midrule
         & & & \textbf{\checkmark} & 75.58 & 2.60 & 66.75 & 2.45 \\ 
         & & \textbf{\checkmark} & \textbf{\checkmark} & 74.31 & 2.89 & 65.95 & 3.60 \\ 
         & \textbf{\checkmark} & \textbf{\checkmark} & \textbf{\checkmark} & 75.90 & 1.36 & 65.73 & 1.96 \\ 
         \textbf{\checkmark} & \textbf{\checkmark} & \textbf{\checkmark} & \textbf{\checkmark} & 76.66 & 2.70 & 67.10 & 2.23 \\  \bottomrule \bottomrule
        \end{tabular}
        \label{tab:5a}
        }
        }
    \end{minipage}
    \begin{minipage}[t]{\columnwidth}
        \centering
        \subfloat[]{
        \resizebox{0.75\columnwidth}{!}{
        \begin{tabular}{cccc|cc|cc}
        \toprule \toprule
         \multicolumn{4}{c|}{Layer} & \multicolumn{2}{c|}{Local Test (\%)} & \multicolumn{2}{c}{Global Test (\%)} \\ \cmidrule{1-8}
                                         1 & 2 & 3 & 4 & Mean & Std & Mean & Std \\  \hline
         \multicolumn{8}{c}{$\alpha$=1} \\ \hline
         & & & \textbf{\checkmark} & 88.66 & 1.11 & 84.01 & 1.26 \\ 
         & & \textbf{\checkmark} & \textbf{\checkmark} & 88.60 & 2.24 & 84.68 & 1.37 \\ 
         & \textbf{\checkmark} & \textbf{\checkmark} & \textbf{\checkmark} & 88.68 & 2.06 & 84.25 & 1.88 \\ 
         \textbf{\checkmark} & \textbf{\checkmark} & \textbf{\checkmark} & \textbf{\checkmark} & 88.83 & 1.85 & 84.26 & 2.13 \\ \hline
        
         \multicolumn{8}{c}{$\alpha$=0.5} \\ \hline
         & & & \textbf{\checkmark} & 87.12 & 1.53 & 83.32 & 1.86 \\ 
         & & \textbf{\checkmark} & \textbf{\checkmark} & 86.86 & 1.96 & 83.41 & 1.49 \\ 
         & \textbf{\checkmark} & \textbf{\checkmark} & \textbf{\checkmark} & 87.11 & 2.08 & 83.87 & 2.08 \\ 
         \textbf{\checkmark} & \textbf{\checkmark} & \textbf{\checkmark} & \textbf{\checkmark} & 87.35 & 1.52 & 83.64 & 1.81 \\ \bottomrule \bottomrule
        \end{tabular}
        \label{tab:5b}
        }
        }
        
    \end{minipage}

    \label{tab:tb5}
\end{table}

%% file: table/tab9.tex
\begin{table}[t]
\centering
\scriptsize
\setlength{\tabcolsep}{4pt}
\caption{Performance of MHFL methods under a label flipping data poisoning attack. The four federated clients are indexed using numerical subscripts. All reported results correspond to cross validation means without standard deviations.}
\label{tab:local_global_results}
\begin{tabular}{l|ccccc|c}
\toprule \toprule
\multirow{2}{*}{Method} 
& \multicolumn{5}{c|}{\textbf{Local Test}} 
& \textbf{Global Test} \\
\cmidrule(lr){2-7}
& $M_{1}$ & $M_{2}$ & $M_{3}$ & $M_{4}$ & Mean & Mean \\
\midrule
FedMRL & 62.63 & 67.97 & 59.01 & 70.64 & 65.06 & 57.38 \\
pFedAFM & 71.62 & 74.42 & 49.38 & 72.11 & 66.88 & 58.94 \\
AlFeCo & 69.06 & 66.98 & 68.29 & 70.56 & 68.72 & 59.10 \\
FedCross$^{\dagger}$ & 68.43 & 66.21 & 70.29 & 70.76 & 68.93 & 61.98 \\
FedSKD & 74.61 & 72.40 & 77.15 & 72.10 & 74.07 & 63.85 \\
\bottomrule \bottomrule
\end{tabular}
\label{tab:tb9}
\end{table}

%% file: table/tab6.tex
\begin{table}[htb!]
    \caption{Ablation study on activated timing for similarity knowledge distillation. (a) Results on the FedASD dataset. (b) Results on the FedSkin datasets.}
    \centering
    \begin{minipage}[t]{\columnwidth}
        \centering
        \subfloat[]{
        \resizebox{0.75\columnwidth}{!}{
        \begin{tabular}{c|cc|cc}
        \toprule \toprule
        \multirow{2}{*}{KD Round (\%)} & \multicolumn{2}{c|}{Local Test (\%)} & \multicolumn{2}{c}{Global Test (\%)} \\ \cmidrule{2-5} 
          & Mean & Std & Mean & Std \\ \midrule
        0 & 76.66 & 2.70 & 67.10 & 2.23 \\ 
        25 & 75.33 & 2.27 & 66.99 & 2.46 \\ 
        50 & 73.75 & 1.65 & 65.66 & 2.12 \\ 
        75 & 73.19 & 2.81 & 65.79 & 2.95 \\  \bottomrule \bottomrule
        \end{tabular}
        \label{tab:6a}
        }
        }
        
    \end{minipage}
    \begin{minipage}[t]{\columnwidth}
        \centering
        \subfloat[]{
        \resizebox{0.75\columnwidth}{!}{
        \begin{tabular}{c|cc|cc}
        \toprule \toprule
         \multirow{2}{*}{KD Round (\%)} & \multicolumn{2}{c|}{Local Test (\%)} & \multicolumn{2}{c}{Global Test (\%)} \\ \cmidrule{2-5} 
         & Mean & Std & Mean & Std \\ \hline
         \multicolumn{5}{c}{$\alpha$=1} \\ \hline
         0 & 88.83 & 1.85 & 84.26 & 2.13 \\ 
         25 & 88.20 & 1.09 & 84.62 & 1.26 \\  
         50 & 87.65 & 1.99 & 84.10 & 1.08 \\ 
         75 & 87.37 & 2.28 & 83.81 & 1.95 \\  \hline

        \multicolumn{5}{c}{$\alpha$=0.5} \\ \hline
         0 & 87.35 & 1.52 & 83.64 & 1.81 \\  
         25 & 86.28 & 1.61 & 83.25 & 1.57 \\  
         50 & 86.12 & 2.32 & 83.17 & 2.06 \\ 
         75 & 85.23 & 2.05 & 82.24 & 2.68 \\  \bottomrule \bottomrule
        \end{tabular}
        \label{tab:6b}
        }
        }
    \end{minipage}

    \label{tab:tb6}
\end{table}

%% file: table/tab7.tex
\begin{table*}[htb!]
\caption{Comparison of fairness performance of FedCross and FedSKD on Sex Attribute on both FedASD dataset (left) and FedSkin dataset (right).}
\centering
\renewcommand{\arraystretch}{1.2}
\resizebox{0.80\textwidth}{!}{
\begin{tabular}{|c|cccc|cccc|cccc|}
\hline \hline
\multirow{2}{*}{\centering \textbf{Method}} & \multicolumn{4}{c|}{\textbf{FedASD}} & \multicolumn{8}{c|}{\textbf{FedSkin}} \\ \cline{2-13} 
 & \multirow{2}{*}{\textbf{Male}$\uparrow$} & \multirow{2}{*}{\textbf{Female}$\uparrow$} & \multirow{2}{*}{\textbf{Avg}$\uparrow$} & \multirow{2}{*}{\textbf{$\Delta\downarrow$}} & \multicolumn{4}{c|}{\textbf{$\alpha$=1}} & \multicolumn{4}{c|}{\textbf{$\alpha$=0.5}} \\ \cline{6-13} 
 &  &  &  &  & \textbf{Male}$\uparrow$ & \textbf{Female}$\uparrow$ & \textbf{Avg}$\uparrow$ & \textbf{$\Delta\downarrow$} & \textbf{Male}$\uparrow$ & \textbf{Female}$\uparrow$ & \textbf{Avg}$\uparrow$ & \textbf{$\Delta\downarrow$} \\ \hline
FedCross & 67.46 & 79.60 & 73.53 & 12.14 & 86.77 & 81.59 & 84.18 & 5.18 & 88.88 & 81.34 & 85.11 & 7.54 \\ \hline
FedSKD & 71.57 & 77.34 & \textbf{74.46} & \textbf{5.77} & 89.79 & 85.26 & \textbf{87.52} & \textbf{4.54} & 88.27 & 86.08 & \textbf{87.17} & \textbf{2.19} \\ \hline\hline
\end{tabular}
}
\label{tab:tb7}
\end{table*}

%% file: content/discussion.tex
\section{Discussion}
The proposed FedSKD framework addresses two fundamental challenges in federated learning for medical imaging: model heterogeneity and the associated risks of model drift and knowledge dilution. By adopting an aggregation-free peer-to-peer (P2P) architecture combined with hierarchical similarity knowledge distillation (SKD), FedSKD provides an effective and scalable solution for heterogeneous clinical environments.

\textit{First}, FedSKD eliminates the reliance on centralized aggregation, a core limitation of existing aggregation-based MHFL methods that introduces substantial server-side computational overhead and performance instability. As described in Section 2, the framework operates entirely without server-side aggregation, enabling direct peer-to-peer model exchange. This design not only reduces infrastructure complexity but also avoids performance degradation caused by gradient conflicts during weight aggregation, as evidenced by the superior performance of aggregation-free baselines (FedCross$^{\dagger}$ and FedCross) over aggregation-dependent methods (FedMRL, pFedAFM, AlFeCo, FedAvg, FedProx, FedBN) in Tables \ref{tab:tb2} and \ref{tab:tb3}. Moreover, the model-agnostic nature of FedSKD allows seamless deployment in both heterogeneous and homogeneous FL settings, making it well suited for real-world clinical applications.

\textit{Second}, FedSKD introduces a multi-dimensional similarity knowledge distillation mechanism that enables semantically rich and bidirectional knowledge transfer across clients. By jointly leveraging batch-wise, pixel/voxel-wise, and region-wise SKD, FedSKD facilitates representation learning under non-IID data distributions common in medical imaging. Fine-grained spatial alignment preserves local diagnostic patterns, batch-level semantic correlation enforces global consistency, and region-wise functional connectivity distillation captures clinically meaningful interdependencies. Through mutual learning, local expertise is shared across institutions while frozen prediction heads prevent catastrophic forgetting, enabling stable cross-client knowledge integration. This mechanism effectively mitigates model drift and knowledge dilution, as evidenced by consistent improvements over FedCross in Tables \ref{tab:tb2}--\ref{tab:tb4}, with the full SKD ensemble achieving the best overall performance. Further results in Figure \ref{fig:fig5} demonstrate robust generalization across 17 ABIDE institutions and on the Derm7pt dataset. Importantly, the proposed distillation strategy is architecture-agnostic and readily extendable to different backbone networks.

Beyond overall performance gains, FedSKD simultaneously improves personalization and cross-institutional generalization under both Local and Global Test settings. Notably, it surpasses the idealized \textit{Centralized} baseline (Tables \ref{tab:tb3} and \ref{tab:tb4}), despite the pronounced non-IID characteristics of medical imaging data arising from heterogeneous acquisition protocols, scanner types, and patient populations. This finding highlights FedSKD’s robustness in mitigating distribution shifts while accommodating model heterogeneity in clinically realistic federated learning scenarios.

In summary, FedSKD extends P2P federated learning to the model-heterogeneous setting, allowing clients to retain personalized models while benefiting from direct peer-to-peer knowledge exchange. Through multi-dimensional similarity knowledge distillation, FedSKD effectively alleviates model drift and knowledge dilution, addressing key limitations of existing MHFL and sequential P2P FL approaches in medical image analysis.

While FedSKD represents a significant advancement in model-heterogeneous federated learning, several limitations merit further investigation. \textit{First}, this study primarily evaluates FedSKD on medical image classification. Although the framework is in principle extensible to other tasks, such as segmentation and detection, adapting the underlying model structures requires further validation. Future work will explore FedSKD across a broader range of medical imaging applications. \textit{Second}, although mutual learning increases computational overhead, FedSKD prioritizes robustness and accuracy in heterogeneous P2P settings, and the observed gains suggest a reasonable trade-off between performance and cost. Future work will investigate more efficient training strategies, including iterative DAM–KTM updates and lightweight distillation schemes, such as attention-based or feature-level distillation \cite{zagoruyko2016paying,yim2017gift}. \textit{Third}, the aggregation-free paradigm requires peer-to-peer exchange of entire models, which may introduce privacy risks, including model inversion, membership inference, or gradient extraction attacks \cite{yuan2024decentralized}. Incorporating client-side protection mechanisms, such as partially homomorphic encryption (e.g., Paillier encryption) \cite{paillier1999public}, could enable secure model sharing without exposing raw parameters. \textit{Fourth}, the current implementation relies on random model transfer paths. Although performance is relatively stable across different paths, optimizing transfer orders may further improve efficiency and robustness. \textit{Fifth}, regarding scalability, while the per-client communication cost remains constant as the number of clients grows, completing a full round-robin circulation requires more training rounds. This limitation could be alleviated by adaptive client selection strategies based on domain similarity, network conditions, or resource availability \cite{yuan2024decentralized,wang2022matcha}. \textit{In addition}, FedSKD is not explicitly tailored for long-tail class distributions; integrating imbalance-aware strategies, such as BalanceFL \cite{shuai2022balancefl} or Ratio Loss \cite{wang2021addressing}, represents a promising direction.

\textit{Furthermore}, while this work evaluates FedSKD primarily against data poisoning attacks, extending it with advanced defense and recovery mechanisms, such as Crab \cite{jiang2025towards} and FedCFB \cite{chen2024credible}, could improve robustness against a broader range of adversarial threats. \textit{Lastly}, we believe FedSKD could be extended to fairness-aware federated learning, for example through client re-weighting or bias mitigation methods such as AFL \cite{mohri2019agnostic} and q-FedAvg \cite{li2019fair}, enabling equitable performance across heterogeneous client populations.

%% file: content/conclusion.tex
\section{Conclusion}
In this paper, we propose FedSKD, a peer-to-peer federated learning framework that addresses model heterogeneity by enabling direct knowledge exchange via round-robin circulation of heterogeneous client models. This approach eliminates server dependency while mitigating performance degradation associated with traditional aggregation-dependent methods. Central to FedSKD is our multi-dimensional similarity knowledge distillation (batch-, pixel/voxel-, and region-wise), which ensures bidirectional, semantically rich knowledge transfer across clients. This mechanism prevents catastrophic knowledge forgetting and model drift through progressive refinement and distribution alignment. Extensive evaluations on FedASD (ASD classification) and FedSkin (skin lesion diagnosis) demonstrate FedSKD’s superior generalization and personalization capabilities, validating its applicability to real-world medical FL scenarios.

%% file: supplementary.tex
\section{Supplementary Material}

\subsection{The workflow of FedSKD}
\begin{algorithm}[]
\caption{FedSKD for Multi-Heterogeneous Federated Learning}
\label{alg:fedskd}
\begin{algorithmic}[1]
\Require Clients $\{1,\dots,N\}$ with datasets $\{D_i\}$, heterogeneous models $\{M_i\}$, rounds $T$, SKD weight $\gamma$
\For{$t=1$ \textbf{to} $T$}
    \State Generate a random permutation $\mathcal{O}^t = \pi([1,\dots,N])$
    \For{each client $i$ \textbf{in parallel}}
        \State Receive KTM $\tilde{M} = M_{\mathcal{O}^t_i}$
        \If{$\mathcal{O}^t_i = i$} 
            \State Update $M_i$ on $D_i$ via supervised loss $\mathcal{L}_{CE}$
        \Else
            \State Freeze KTM classifier; update KTM feature extractor only
            \State Perform bidirectional SKD between DAM $M_i$ and KTM $\tilde{M}$
            \Statex\hspace{\algorithmicindent}
            $\mathcal{L}= \mathcal{L}_{CE}(M_i(x_i),y_i)
            +\gamma \mathcal{L}_{SKD}(M_i(x_i),\tilde{M}(x_i))
            +\mathcal{L}_{CE}(\tilde{M}(x_i),y_i)$
            \State Update $M_i$ and $\tilde{M}$ using $\mathcal{L}$
        \EndIf
        \State Retain updated $M_i$
    \EndFor
\EndFor
\State \Return Final heterogeneous models $\{M_i\}$
\end{algorithmic}
\end{algorithm}

\subsection{Additional Experiment on Nature Image Domain}
To further evaluate the generalizability of FedSKD beyond the medical imaging domain, we conducted additional experiments on a natural image dataset.

\input{table/tab-cifar}
\subsection{Datasets}

We benchmark FedSKD against state-of-the-art baselines using CIFAR-10, which consists of 60,000 color images (32×32 pixels) across 10 categories, with 6,000 images per class (50,000 for training and 10,000 for testing). From each class, we randomly sampled 2,000 images, yielding a total of 20,000 samples to construct our federated dataset.

To simulate a challenging non-IID setting, we employed a Dirichlet distribution $Dir(\alpha)$ to control the allocation of class samples across clients. Specifically, we set $\alpha=0.5$ and partitioned the data among four federated clients, resulting in a dataset with pronounced non-IID characteristics, which we refer to as FedCIFAR-10.

\subsubsection{Experimental Results}
Table \ref{tab:cifar} reports the performance of FedSKD compared with state-of-the-art methods under both model-homogeneous and model-heterogeneous settings on FedCIFAR-10. Evaluations were performed using both Local Test and Global Test protocols for the 10-class classification task.

In the model-heterogeneous setting, FedSKD achieves the highest global average performance across both test scenarios, with mean AUC values of 98.51\% (Local Test) and 98.14\% (Global Test). This represents improvements of 0.44\% and 0.43\% over FedCross$^{\dagger}$, the strongest heterogeneous FL baseline. Moreover, FedSKD substantially outperforms three recent MHFL methods: by 4.85\% and 6.62\% against pFedAFM, by 7.16\% and 9.63\% against FedMRL, and by 4.23\% and 6.2\% against AlFeCo under Local and Global Tests, respectively. FedSKD also surpasses the lower-bound Local training baseline by 4.36\% and 6.43\%, demonstrating its strong capability to adapt to heterogeneous data. Similar superiority is observed in model-homogeneous settings, where FedSKD consistently outperforms all homogeneous baselines.

Taken together, these results confirm that FedSKD consistently outperforms state-of-the-art baselines under both homogeneous and heterogeneous scenarios. This demonstrates that our framework is not limited to medical applications but also generalizes effectively to non-medical datasets. At the same time, the benefits of FedSKD are particularly pronounced in medical imaging tasks, where data heterogeneity is more severe, underscoring its practical value in real-world clinical settings. Furthermore, aggregation-free peer-to-peer approaches (FedSKD, FedCross variants) consistently outperform aggregation-based methods (FedMRL, pFedAFM, AlFeCo, FedAvg, FedProx, FedBN), highlighting the inherent limitations of parameter averaging in handling data heterogeneity.

%% file: table/tab-cifar.tex
\begin{table*}[htb!]
\caption{Performance comparison across methods in the local and global test scenarios on the FedCIFAR10 dataset. The best results under the model-heterogeneous setting are highlighted in bold. The symbol $^{*}$ indicates that the performance difference between our FedSKD and the second-best baseline methods under the same FL setting is statistically significant at $p<0.05$.}
\centering
\resizebox{\textwidth}{!}{
\begin{tabular}{c|ccccc|ccccc}
\hline
\hline
Method & \multicolumn{5}{c|}{Local Test} & \multicolumn{5}{c}{Global Test} \\ \cline{2-11} 
(FedASD) & $M_{1}$ & $M_{2}$ & $M_{3}$ & $M_{4}$ & Mean & $M_{1}$ & $M_{2}$ & $M_{3}$ & $M_{4}$ & Mean \\ \hline
\multicolumn{1}{c|}{} & \multicolumn{10}{c}{\textbf{Homogeneous Models}} \\ \hline
Centralized & 98.50$\pm$0.18  & 99.07$\pm$0.32  & 98.58$\pm$0.22  & 98.96$\pm$0.27  & 98.78$\pm$0.10  & \textbackslash  & \textbackslash  & \textbackslash  & \textbackslash  & 98.69$\pm$0.09 \\ \hline

Local  & 91.01$\pm$1.19  & 94.96$\pm$1.01  & 95.51$\pm$0.41  & 94.78$\pm$0.96  & 94.07$\pm$0.43  & 88.98$\pm$0.36  & 91.28$\pm$0.55  & 94.82$\pm$0.40  & 91.87$\pm$0.45  & 91.74$\pm$0.13 \\ \hline

FedAvg & 97.04$\pm$0.23  & 98.64$\pm$0.28  & 97.44$\pm$0.24  & 98.49$\pm$0.33  & 97.90$\pm$0.05  & \textbackslash  & \textbackslash  & \textbackslash  & \textbackslash  & 97.77$\pm$0.10 \\ \hline

FedProx & 96.75$\pm$0.59  & 98.62$\pm$0.24  & 97.49$\pm$0.15  & 98.42$\pm$0.30  & 97.82$\pm$0.18  & \textbackslash  & \textbackslash  & \textbackslash  & \textbackslash  & 97.72$\pm$0.22 \\ \hline

FedBN & 97.36$\pm$0.23  & 98.33$\pm$0.38  & 97.80$\pm$0.44  & 98.38$\pm$0.57  & 97.97$\pm$0.24  & 97.87$\pm$0.25  & 97.93$\pm$0.18  & 97.98$\pm$0.26  & 97.89$\pm$0.18  & 97.92$\pm$0.21 \\ \hline

FedMRL  & 90.74$\pm$0.16  & 94.25$\pm$1.76  & 93.90$\pm$0.16  & 93.45$\pm$0.19  & 93.08$\pm$0.15  & 87.76$\pm$1.38  & 90.27$\pm$1.56  & 93.29$\pm$1.82  & 90.53$\pm$1.40  & 90.46$\pm$1.44 \\ \hline

pFedAFM  & 90.80$\pm$1.66  & 94.55$\pm$0.52  & 94.92$\pm$0.28  & 94.41$\pm$0.98  & 93.67$\pm$0.46  & 88.97$\pm$0.43  & 91.37$\pm$0.41  & 94.28$\pm$0.58  & 91.80$\pm$0.32  & 91.61$\pm$0.21 \\ \hline

AlFeCo  & 91.60$\pm$1.95  & 95.70$\pm$0.76  & 95.81$\pm$0.32  & 95.10$\pm$1.05  & 94.55$\pm$0.53  & 89.09$\pm$0.98  & 91.61$\pm$0.46  & 95.20$\pm$0.38  & 92.19$\pm$0.42  & 92.02$\pm$0.32 \\ \hline

FedCross & 97.54$\pm$0.18 & 98.59$\pm$0.29 & 98.10$\pm$0.34 & 98.36$\pm$0.37 & 98.15$\pm$0.14 & \textbackslash & \textbackslash & \textbackslash & \textbackslash & 97.79$\pm$0.34 \\ \hline

\rowcolor{gray!20} FedSKD & 97.99$\pm$0.33  & 98.95$\pm$0.25  & 98.62$\pm$0.30  & 98.74$\pm$0.42  & \textbf{98.58$\pm$0.16}  & 98.10$\pm$0.28  & 98.19$\pm$0.22  & 98.29$\pm$0.21  & 98.19$\pm$0.21  & \textbf{98.19$\pm$0.22} \\ \hline
\multicolumn{1}{c|}{} & \multicolumn{10}{c}{\textbf{Heterogeneous Models}} \\ \hline

Centralized & 98.48$\pm$0.18  & 99.03$\pm$0.32  & 98.50$\pm$0.22  & 98.89$\pm$0.38  & 98.73$\pm$0.10  & 98.73$\pm$0.04  & 98.70$\pm$0.10  & 98.64$\pm$0.12  & 98.59$\pm$0.14  & 98.66$\pm$0.09 \\ \hline

Local  & 91.67$\pm$0.14  & 95.15$\pm$0.91  & 95.11$\pm$0.13  & 94.67$\pm$0.91  & 94.15$\pm$0.36  & 89.29$\pm$0.55  & 91.64$\pm$0.57  & 94.60$\pm$0.42  & 91.30$\pm$0.71  & 91.71$\pm$0.34 \\ \hline

FedMRL  & 89.50$\pm$1.86  & 92.45$\pm$1.20  & 92.01$\pm$1.74  & 91.42$\pm$1.73  & 91.35$\pm$1.14  & 85.99$\pm$1.32  & 88.38$\pm$1.49  & 91.34$\pm$1.66  & 88.34$\pm$1.71  & 88.51$\pm$1.35 \\ \hline

pFedAFM  & 90.73$\pm$2.30  & 94.87$\pm$0.73  & 94.89$\pm$0.23  & 94.14$\pm$0.72  & 93.66$\pm$0.67  & 88.91$\pm$1.07  & 91.49$\pm$0.42  & 94.14$\pm$0.48  & 91.54$\pm$0.25  & 91.52$\pm$0.44 \\ \hline

AlFeCo  & 91.16$\pm$2.16  & 95.06$\pm$1.08  & 95.75$\pm$0.26  & 95.13$\pm$0.82  & 94.28$\pm$0.73  & 89.20$\pm$0.77  & 91.58$\pm$0.60  & 95.01$\pm$0.47  & 91.98$\pm$0.40  & 91.94$\pm$0.30 \\ \hline

FedCross$^{\dagger}$ & 97.66$\pm$0.32  & 98.33$\pm$0.25  & 98.02$\pm$0.27  & 98.29$\pm$0.37  & 98.07$\pm$0.10  & 97.86$\pm$0.21  & 97.64$\pm$0.34  & 97.67$\pm$0.25  & 97.65$\pm$0.14  & 97.71$\pm$0.23 \\ \hline

\rowcolor{gray!20} FedSKD & 97.89$\pm$0.12  & 99.04$\pm$0.17  & 98.50$\pm$0.24  & 98.62$\pm$0.34  & \textbf{98.51$\pm$0.15}$^{*}$  & 98.06$\pm$0.27  & 98.16$\pm$0.30  & 98.21$\pm$0.25  & 98.12$\pm$0.30  & \textbf{98.14$\pm$0.27}$^{*}$ \\ \hline \hline
\end{tabular}
}
\label{tab:cifar}
\end{table*}